\pdfoutput=1

\documentclass{article} %
\usepackage{iclr2027_conference,times}

\usepackage{amsmath,amsfonts,bm}

\def\eqref#1{equation~\ref{#1}}

\def\1{\bm{1}}

\DeclareMathAlphabet{\mathsfit}{\encodingdefault}{\sfdefault}{m}{sl}
\SetMathAlphabet{\mathsfit}{bold}{\encodingdefault}{\sfdefault}{bx}{n}

  \usepackage{graphicx}
  \usepackage{booktabs}
  \usepackage{subcaption}
  \usepackage{amsmath}
  \usepackage{multirow}
  \usepackage{comment}
  \usepackage{longtable}
\usepackage{hyperref}
\usepackage{url}
\usepackage{xspace}

\title{Cross-Prompt Inference Mitigates Rare-Class Suppression in Imbalanced VLM Finetuning}

\author{Emanuel Ben Baruch, Sivan Biham, Alon Shoshan, Yochai Zvik, Avi Ben-Cohen \\
Amazon \\
\texttt{\{emanbb, sbiham, alonshos, yochai, avibc\}@amazon.com}}

\makeatletter
\DeclareRobustCommand\onedot{\futurelet\@let@token\@onedot}
\def\@onedot{\ifx\@let@token.\else.\null\fi\xspace}

\def\ie{\emph{i.e}\onedot}

\makeatother

\iclrfinalcopy
\begin{document}

\maketitle

\begin{abstract}
When a vision-language model is finetuned on class-imbalanced data, rare-class recall can fall well below the model's own zero-shot performance.
In this regime, we uncover a surprising matched-prompt penalty whereby the standard choice of reusing the descriptive finetuning prompt at inference can yield lower rare-class recall than using a different descriptive prompt.
We call this inference-time strategy cross-prompt inference.
We evaluate diverse VLMs across six datasets and find that cross-prompt inference consistently recovers rare-class recall, with varying trade-offs in frequent-class recall.
The effect reflects a train-eval prompt interaction rather than prompt quality alone, and does not
arise with non-descriptive prompt changes. It is strongest when imbalance and
visual confusability produce substantial rare-class suppression, and is weak or
absent when such suppression does not occur. Using Tuning Contribution (TuCo), which decomposes a model's computation into pretrained and finetuning parts,
we find that cross-prompt inference reduces reliance on finetuning-induced
computation and shifts the model toward its pretrained component. Reducing that
finetuning contribution also substantially reduces the cross-prompt advantage.
These results show that rare-class knowledge suppressed by imbalanced finetuning can remain accessible under a different inference prompt.
\end{abstract}
    
\section{introduction}

Vision-language models are increasingly adapted to downstream image classification through supervised finetuning on labeled data~\citep{zhang2024visuallygrounded,geigle2024african}. In many
applications, this data is long-tailed. A small set of frequent classes has abundant supervision, while rare classes have only a few examples. 
Such imbalance can bias a supervised classifier toward frequent classes and degrade performance where supervision is scarce~\citep{cui2019classbalanced,cao2019learning,kang2020decoupling,menon2021longtail}.

VLMs offer an additional degree of freedom. Their label space can be expressed through natural language, either with class names alone or with short descriptions of visual attributes and contrastive cues. These
descriptions can surface class knowledge acquired during pretraining. This may be particularly useful when rare classes have little downstream supervision.

\begin{figure*}[t]
      \centering
      \captionsetup[subfigure]{
          font=small,
          justification=centering,
          skip=1pt
      }

      \begin{subfigure}[t]{0.60\textwidth}
          \centering
          \begin{minipage}[c][0.27\textheight][c]{\linewidth}
              \centering
              \raisebox{0.6cm}{%
                  \includegraphics[
                      width=0.98\linewidth,
                      height=0.27\textheight,
                      keepaspectratio
                  ]{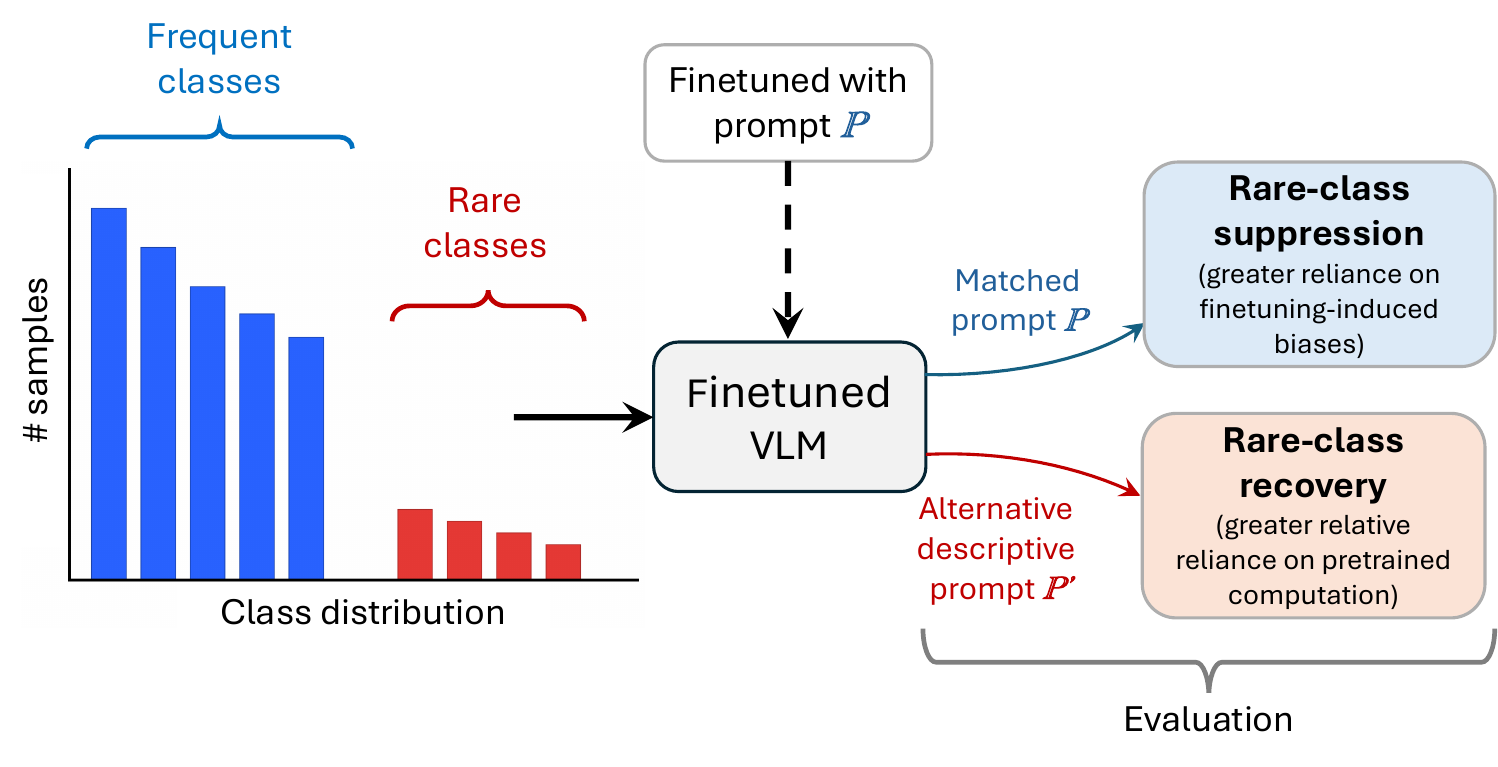}%

              }
          \end{minipage}
          \vspace{-0.75cm}
          \caption{Cross-prompt inference}
          \label{fig:cross-prompt-overview}
      \end{subfigure}\hfill
      \begin{subfigure}[t]{0.36\textwidth}
          \centering
          \begin{minipage}[c][0.27\textheight][c]{\linewidth}
              \centering
              \hspace*{-0.25cm}%
              \raisebox{-0.45cm}{%
                  \includegraphics[
                      width=0.98\linewidth,
                      height=0.27\textheight,
                      keepaspectratio
                  ]{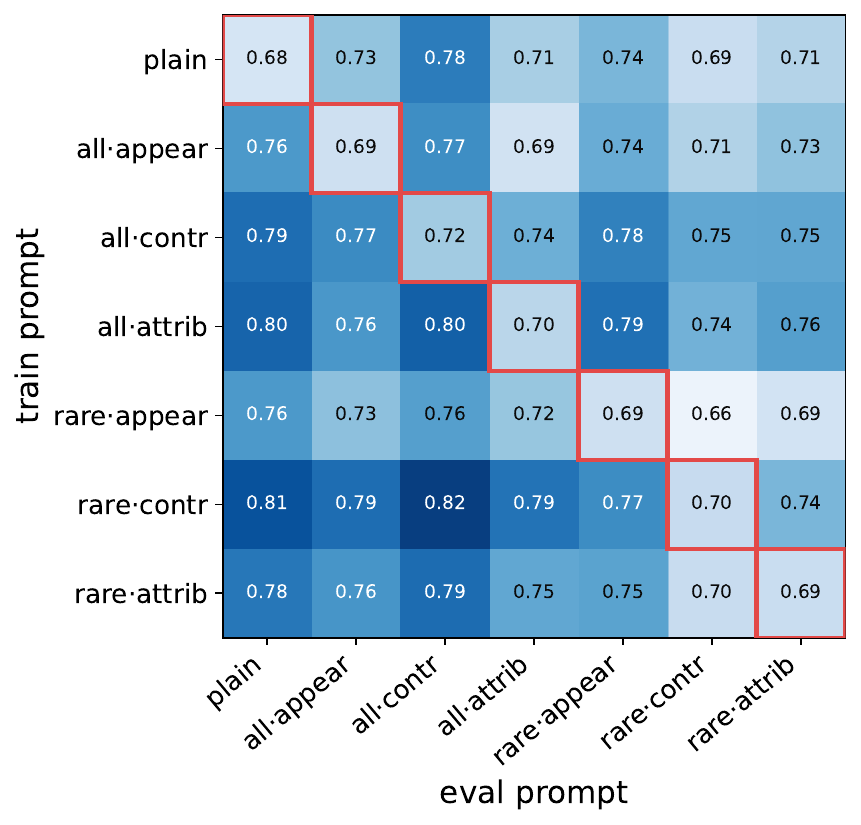}%
              }
          \end{minipage}
          \vspace{-0.75cm}
          \caption{Food-101 matrix}
          \label{fig:fig1_heatmap}
      \end{subfigure}

\caption{\textbf{The matched-prompt penalty under class-imbalanced VLM finetuning.}
(a) Finetuning with prompt $P$ on an imbalanced distribution favors frequent classes
and suppresses rare ones. Reusing $P$ at inference preserves this suppression, while
a different descriptive prompt $P'$ mitigates it without retraining by enabling the
model to draw less on the computation introduced by finetuning.
(b) A $7\times7$ Food-101 cross-prompt matrix, in which the matched cell is never
its row maximum.}
      \label{fig:fig1}
  \end{figure*}

We find that severe imbalance can suppress this rare class knowledge during finetuning. Across fine-grained benchmarks, rare class recall can fall well below the model's zero-shot level while performance on frequent classes improves.
Evaluating with descriptive prompts, we uncover a surprising interaction between the finetuning and inference prompts.
When finetuning uses a descriptive prompt $P$, evaluating with that
same prompt can yield lower rare-class recall than evaluating the identical finetuned model with a different descriptive prompt $P'$.
This switch can recover substantial rare-class recall, with varying trade-offs in frequent-class recall.
We call this effect the \emph{matched-prompt penalty} and the inference-time strategy \emph{cross-prompt inference} (see Figure~\ref{fig:fig1}).

The matched-prompt penalty reflects an interaction between the training and evaluation prompts rather than the quality of either prompt alone. Moreover, changing between non-descriptive prompts does not recover rare-class recall, indicating that descriptive content  is necessary.

To better understand why cross-prompt inference recovers rare classes, we analyze how the evaluation prompt changes the relative influence of pretrained and finetuning-induced computation. Using TuCo~\citep{nuti2025tuco}, we find that
mismatched descriptive prompts rely less on computation introduced during finetuning.
Reducing that contribution also substantially reduces the cross-prompt advantage.
Together, these results support the view that cross-prompt inference shifts computation away from imbalanced finetuning and toward the pretrained component, recovering rare-class knowledge suppressed by finetuning.

We further characterize when this recovery occurs. Rare class suppression grows with both the imbalance ratio and the visual similarity between frequent and rare classes.
Cross-prompt inference therefore helps most in fine-grained and highly imbalanced settings, where finetuning has substantially degraded rare-class performance.
When rare classes remain intact after finetuning, the matched-prompt penalty is weak or absent.
We further find that cross-prompt inference complements standard oversampling, yielding further gains in rare-class recall.
We evaluate these findings across six datasets and six VLMs spanning several model families and scales.

Our contributions are threefold. First, we identify the matched-prompt penalty in imbalanced VLM finetuning and
show that cross-prompt inference can mitigate the resulting rare-class suppression.
Second, TuCo analysis shows that cross-prompt inference reduces reliance on finetuning-induced computation, and that its recovery
advantage shrinks when this contribution is removed. 
Third, we characterize the conditions under which the effect arises. The penalty strengthens with imbalance severity and class similarity, persists after oversampling, and disappears without descriptive prompts.

\section{related work}

\noindent\textbf{Long-tailed recognition.}
Class imbalance is commonly addressed through re-sampling or loss re-weighting by class frequency~\citep{buda2018systematic,cui2019classbalanced,cao2019learning}.
\citet{menon2021longtail} instead adjust the decision rule at inference by correcting for the label prior, building on evidence that imbalance affects the classifier more than the representation~\citep{kang2020decoupling}.
Recent work extends long-tailed learning to VLMs. Methods adapt CLIP to long-tailed data~\citep{tian2022vlltr,dong2023lpt}, show that heavy fine-tuning can hurt tail classes~\citep{shi2024lift}, remove pretraining label bias~\citep{zhu2023gla}, rebalance VLM training data~\citep{song2025adr}, and document underrepresentation of rare concepts in pretraining~\citep{parashar2024neglected}. These methods modify training or output logits, whereas our approach changes only the inference prompt.

\noindent\textbf{Class descriptions for VLM classification.}
CLIP classifies images by matching them to text prompts~\citep{radford2021learning}, while instruction-tuned VLMs can be fine-tuned to emit class names directly~\citep{liu2023llava}. Many works improve classification by describing classes rather than naming them. \citet{menon2023description} and \citet{pratt2023cupl} generate class descriptions with an LLM, while \citet{esfandiarpoor2024fudd} and \citet{lee2025enhancing} describe differences between visually similar classes, inspiring our contrastive style. These gains may not depend entirely on description semantics, as \citet{roth2023waffling} find random words can perform comparably and \citet{ma2025descsemantics} attribute part of the benefit to semantic-agnostic ensembling. Prior description-based methods use frozen encoder, while prompt adaptation either learns prompts during training~\citep{zhou2022coop} or optimizes them at test time on an unchanged model~\citep{shu2022tpt}. Prompt wording and format can also strongly affect accuracy~\citep{sclar2024quantifying}. We instead use descriptive prompts during supervised fine-tuning and study the effect of changing them at inference.

\noindent\textbf{Forgetting in VLM fine-tuning.}
Fine-tuning can degrade pretrained abilities, including CLIP zero-shot robustness~\citep{wortsman2022wiseft,kumar2022finetuning}, and MLLMs can even underperform their own vision encoder~\citep{zhai2024investigating}.
Several works interpret this as a behavioral shift rather than knowledge loss.
\citet{zheng2025spurious} attribute it to reduced task alignment, while \citet{kotha2024understanding} show that pretrained capabilities can be recovered by moving
prompts away from the fine-tuning distribution, mainly by translating English prompts into other languages.
Our setting differs in that the task and label space remain fixed and the degradation is concentrated on rare classes under class imbalance.

Prior work finds that mismatching the fine-tuning and inference prompts is harmful. Unseen prompt formulations can reduce performance~\citep{aissi2025reinforcement}, and prompt-format consistency is often considered important for reliable transfer~\citep{yun2025price}. We identify class imbalance as a regime where this behavior changes for rare classes, with the matched descriptive prompt becoming suboptimal and an alternative descriptive prompt recovering part of the suppressed rare-class performance.

\section{Problem Setting and Cross-Prompt Inference}

We consider finetuning a vision-language model (VLM) $f$ for image classification
under class imbalance. Let $\mathcal{C}=\{c_1,\ldots,c_K\}$ be the set of class names,
and let $P_t$ denote the textual prompt used during finetuning. The prompt presents
the classification task and may additionally include a description of each class.
We finetune on a training set of image-label pairs
$\{(x_i,y_i)\}_{i=1}^{N}$ with $y_i\in\mathcal{C}$. The training distribution is
class-imbalanced, ranging from frequent classes with abundant examples to rare
classes with very few. Let $n_c$ be the number of training examples for class $c$.
We characterize the imbalance by the ratio $\rho = \max_c n_c / \min_c n_c \gg 1$.

For each example, the model is conditioned on $P_t$ and the image $x_i$, and is
trained to generate the target class name $y_i$ by minimizing the standard
autoregressive cross-entropy over the response tokens. We denote the resulting
parameters by $\theta_{P_t}$.

\subsection{Cross-Prompt Inference}

At inference, let $P_e$ denote the evaluation prompt.
The conventional choice is to reuse the prompt seen during finetuning, giving the
matched configuration $P_e=P_t$. This choice is natural because inference is then
aligned with the finetuning condition and is consistent with prior work showing that
unseen prompt formulations can degrade performance and that finetuning may
overfit to the prompt used during training~\citep{sun2024evaluating,aissi2025reinforcement}.

Under severe class imbalance, however, we find that this matched configuration can
be suboptimal for rare classes. Rare-class performance can be substantially
degraded by finetuning, sometimes falling below the model's own zero-shot
performance. Our hypothesis is that, under imbalanced finetuning, the descriptive
training prompt becomes coupled to a prediction bias toward frequent classes.
Reusing the same prompt at inference may preserve this coupling and suppress rare
classes, even when the prompt contains useful descriptive information. Evaluating
with a different descriptive prompt may weaken this prompt-specific bias and allow
the model to rely more on pretrained class knowledge.
This motivates cross-prompt inference, where the finetuned model is
evaluated under a different descriptive prompt,
\begin{equation}
\hat{y} = f(x,P_e;\theta_{P_t}),
\qquad P_e \neq P_t.
\label{eq:cross_prompt}
\end{equation}

Cross-prompt inference changes neither the finetuned parameters nor the decoding
procedure. It changes only the prompt used at inference. Importantly, the effect
is not produced by prompt mismatch alone. As shown in section ~\ref{sec:non_descriptive}, changing
between description-free prompts does not yield the same rare-class recovery.

\subsection{Quantifying the Matched-Prompt Penalty}
Given a set of prompts $\mathcal{P}=\{P_1,\ldots,P_m\}$, we summarize the interaction
between the training and evaluation prompts with a matrix
\begin{equation}
M_{ij}
=
\operatorname{Recall}_{\mathrm{rare}}
\bigl(f(x,P_j;\theta_{P_i})\bigr).
\label{eq:matrix}
\end{equation}
The diagonal entries $M_{ii}$ correspond to the matched prompt, while the off-diagonal
entries correspond to cross-prompt inference (see Figure~\ref{fig:fig1_heatmap} for an example).

For a model finetuned with $P_i$, we measure the recovery from changing the
evaluation prompt with the average mismatch gain and the best mismatch gain
\begin{equation}
\Delta_i^{\mathrm{avg}}
=
\frac{1}{m-1}\sum_{j\neq i} M_{ij}
-
M_{ii},
\qquad
\Delta_i^{\mathrm{best}}
=
\max_{j\neq i} M_{ij}
-
M_{ii},
\label{eq:gains}
\end{equation}
and we report both averaged over the training prompts. 
The average mismatch gain measures whether the benefit extends across alternative
evaluation prompts, whereas the best mismatch gain captures the largest recovery
within the tested prompt set. Together, they characterize the average and
potential benefit of cross-prompt inference.

\subsection{Finetuning Contribution Analysis with TuCo}
\label{sec:tuco}

Our hypothesis is that cross-prompt inference helps because a mismatched
descriptive prompt reduces the influence of the computation introduced by
imbalanced finetuning, allowing the pretrained component to play a larger role.
To test this hypothesis, we adopt Tuning Contribution (TuCo)~\citep{nuti2025tuco}, which splits the
update each transformer layer adds to the residual stream into the update the pretrained layer would have made, $\mathrm{PTC}_{\ell}$, and the extra update
from finetuning, $\mathrm{FTC}_{\ell}$, so that
$h_{\ell+1} = h_{\ell} + \mathrm{PTC}_{\ell} + \mathrm{FTC}_{\ell}$ exactly. The
TuCo score is the relative magnitude of the accumulated finetuning updates at the
final prompt token, between $0$ and $1$. 
This internal view enables us to detect changes that cancel out by the final
layer and leave the output unchanged. More importantly, it lets us rescale the finetuning contribution rather than only compare the two models.

\citet{nuti2025tuco} showed that successful jailbreaks circumvent safety
finetuning while exhibiting a lower finetuning contribution. We test the
analogous pattern for imbalance-induced bias, extending their text-only
decomposition to a vision-language model.

Our main analysis rescales the finetuning component by a factor $\lambda$,
\begin{equation}
    h_{\ell+1}
    =
    h_{\ell}
    +
    \mathrm{PTC}_{\ell}
    +
    \lambda\,\mathrm{FTC}_{\ell},
    \qquad \lambda \in [0,1].
    \label{eq:tuco_lambda}
\end{equation}
At $\lambda=1$, the intervention reproduces the finetuned decoder computation
exactly. At $\lambda=0$, it removes the TuCo finetuning contribution from each
decoder-layer update, which lets us control how strongly the finetuned and pretrained
components contribute.
If the cross-prompt advantage
comes from attenuating imbalanced finetuning, it should shrink as $\lambda$ decreases.

We report the TuCo score, which measures how strongly each evaluation prompt engages the finetuning component.
These analyses test whether cross-prompt inference shifts computation from the imbalanced finetuning toward the pretrained component, recovering rare-class predictions that the matched prompt suppresses.
Results are in Section~\ref{sec:tuco_results} and implementation details in Appendix~\ref{app:tuco}.

\section{Experiments}
\label{sec:experiments}

\begin{figure}[t]
    \centering
    \begin{minipage}[t]{0.55\linewidth}
        \vspace{0pt}
        \centering
        \begin{subfigure}[b]{0.49\linewidth}
            \includegraphics[width=\linewidth]{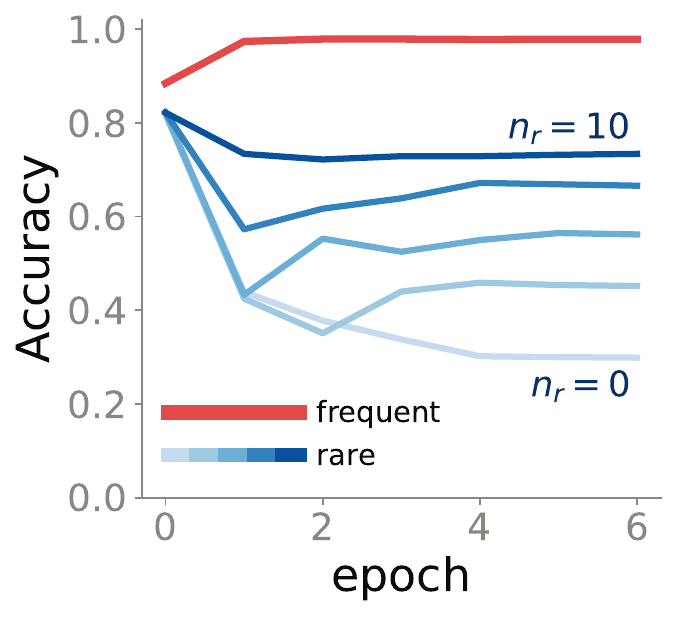}
            \vspace{-3mm}
            \caption{Food-101}
            \label{fig:forget-food}
        \end{subfigure}
        \hfill
        \begin{subfigure}[b]{0.49\linewidth}
            \includegraphics[width=\linewidth]{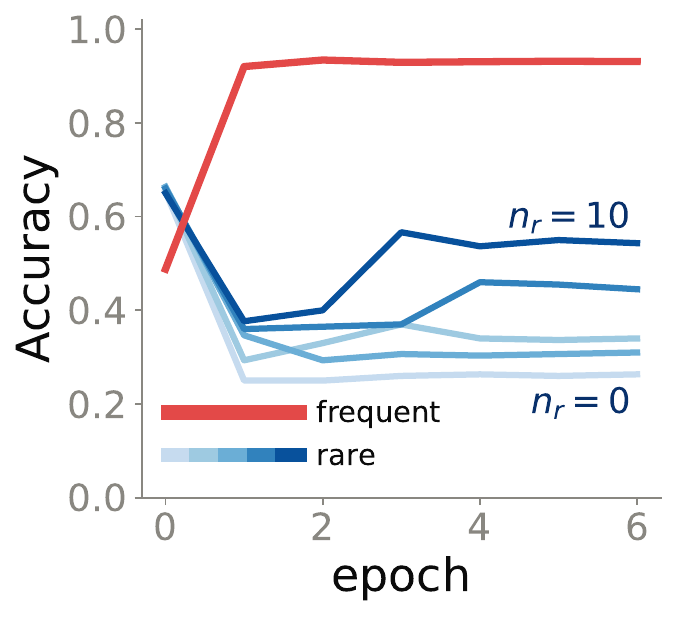}
            \vspace{-3mm}
            \caption{FGVC-Aircraft}
            \label{fig:forget-aircraft}
        \end{subfigure}
    \end{minipage}
    \hfill
    \begin{minipage}[t]{0.44\linewidth}
        \vspace{0pt}
        \vspace{-\abovecaptionskip}
        \caption{\textbf{Rare-class suppression.}
         Accuracy vs.\ epoch for frequent classes (red) and rare classes with $n_r$
        samples per class (blue),
        where $n_r\!=\!0$ trains on frequent classes only. Epoch $0$ is the zero-shot
        model. Frequent accuracy saturates near ceiling while rare accuracy drops
        below its zero-shot value, more so as $n_r$ decreases. The remaining datasets
        show the same pattern (Appendix~\ref{app:suppression}).}
        \label{fig:rare_suppression}
    \end{minipage}
\end{figure}

\providecommand{\PH}[1]{\textcolor{red}{\textbf{[#1]}}}

\subsection{Experimental Setting}
\label{sec:exp_setting}
\noindent\textbf{Datasets.}
We use six datasets: Food-101~\citep{food101}, CUB-200-2011~\citep{cub200}, FGVC-Aircraft~\citep{fgvcaircraft}, Stanford Dogs~\citep{stanforddogs}, PlantNet-300K~\citep{plantnet}, and ImageNet~\citep{russakovsky2015imagenet}.
To study cross-prompt inference across different imbalance settings, we construct controlled subsets by splitting each dataset into 10 frequent and 10 rare classes, so that per-class descriptions remain practical to include in the VLM prompt.

\noindent\textbf{Models.}
We evaluate six VLMs spanning multiple families and a range of sizes, including the Qwen3-VL scale ladder at 2B, 4B, and 8B~\citep{qwen3vl}, Qwen2.5-VL-3B~\citep{qwen25vl}, Gemma-3-4B~\citep{gemma3}, and InternVL-3.5-2B~\citep{internvl35}.

\noindent\textbf{Descriptive prompts.}
We experiment with seven prompts, constructed from three description styles and a choice of whether to describe all classes or only the rare ones. We attach a short description to each class in one of three styles.
These include an appearance style that gives a plain visual description, a contrastive style that names the feature distinguishing a class from its most confusable neighbor, and an attribute style that lists the class's key attributes in a structured form. Each style is applied either to all classes or to the rare classes only, giving six descriptive prompts.
Together with the plain, description-free prompt, this yields the seven prompts used for both finetuning and inference.
For each model type, we finetune seven models, one for each of the seven prompts.
We then evaluate each finetuned model under all seven prompts, yielding a $7 \times 7$ train-evaluation matrix (see the right panel of Figure~\ref{fig:fig1_heatmap} for an example on Food-101).
The full prompts for each dataset are provided in Appendix~\ref{app:prompts}.

\subsection{Rare-class suppression}
\label{sec:suppression}
We begin by examining the impact of high class imbalance on VLM image classification.
For a given dataset we finetune Qwen3-VL-2B under the plain prompt and track frequent- and rare-class accuracy over training epochs.
We vary the number of training samples per rare class $n_r$, which sets the imbalance severity (Figure~\ref{fig:rare_suppression} and Appendix~\ref{app:suppression}).

To construct each imbalanced dataset, we group classes into confusable families by
visual similarity, using $14$ visually similar classes and $6$ distinct anchors per
dataset. This mix spans both regimes within a dataset, since look-alike pairs are
where imbalance can suppress a class while distinct anchors serve as a within-dataset
control. Within each pair of visually similar classes, one class is designated
frequent ($n_f$ images per class) and its look-alike rare ($n_r \ll n_f$). Distinct
anchor classes (those without a close look-alike) are divided evenly. 
Each experiment is repeated with three random seeds.
The seed determines the subset of images drawn for each rare class, as well as model initialization and data ordering. 
The class pairs, the frequent/rare assignment, and the frequent-class training images are identical.

Across the fine-grained datasets, frequent-class accuracy rises quickly and saturates
near $100\%$ within the first epoch, while rare-class accuracy drops and remains
suppressed for the rest of the training. 
The suppression deepens as the imbalance becomes
more severe, with rare accuracy falling progressively lower as $n_r$ decreases.

\begin{figure*}[t]
  \centering
  \includegraphics[width=\linewidth]{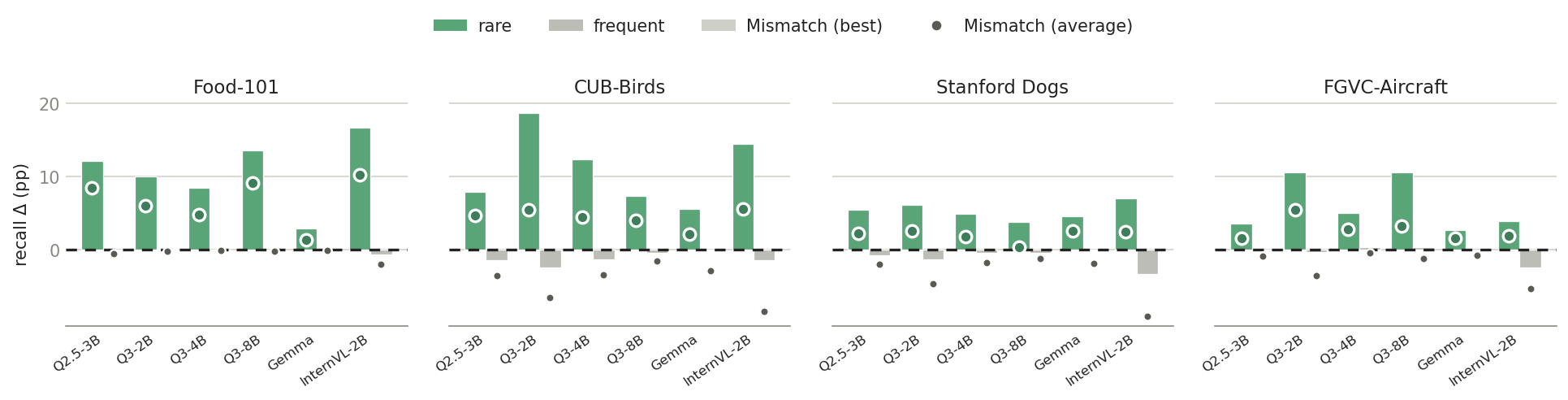}
\caption{\textbf{Cross-prompt recovery of rare recall across models}. Each panel corresponds
to one dataset and shows, for each model, the recall change relative to matched for rare
(green) and frequent (grey) classes. The dot is the averaged mismatch and the
bar is the best mismatch. The rare gain is positive on every model and dataset,
with a trade-off in frequent-class recall.}
  \label{fig:rescue}
\end{figure*}

\subsection{Matched-Prompt Penalty and Cross-Prompt Inference}
\label{sec:main-result}

In this section, we present our main results, showing that finetuning a VLM on class-imbalanced
data with a descriptive prompt makes that same prompt a poor choice at inference for
the rare classes, while a different descriptive prompt (cross-prompt inference)
improves rare-class recall.

Figure~\ref{fig:fig1_heatmap} demonstrates the effect on Food-101 as a $7\times7$
matrix whose rows are training prompts and columns are evaluation prompts. 
For example, the rows trained with the descriptive-all-attribute or
descriptive-rare-contrastive prompts stand out, reaching rare recall of $0.80$ and
$0.82$ under other evaluation prompts, yet falling to among their lowest values
($0.69$ and $0.70$) exactly when evaluated with the prompt they were trained on.
We define this gap, between a model's matched
cell and its improved mismatched cells, as the matched-prompt penalty.

The penalty does not simply reflect some prompts being weak, \ie, no descriptive prompt is bad to train with or to evaluate with on its own.
Rare recall drops mainly when the
training and evaluation prompts coincide, making this a train-eval interaction rather than an effect of either prompt alone.

\begin{table*}[t]
  \centering
\caption{\textbf{Rare and overall recall (\%).}
Matched, average mismatched, and best mismatched results are shown,
with $\Delta$ vs.\ matched in parentheses.
ZS is zero-shot overall recall averaged over the seven prompts.
Cross-prompt inference recovers rare recall under imbalanced
and oversampled training.}
  \label{tab:macrofull}
  \small
  \setlength{\tabcolsep}{3pt}
  \renewcommand{\arraystretch}{1.05}
  \begin{tabular}{@{}ll ccc cccc@{}}
    \toprule
    & & \multicolumn{3}{c}{\textbf{Rare} {\scriptsize($\Delta$ vs.\ matched)}}
        & \multicolumn{4}{c}{\textbf{Overall}} \\
    \cmidrule(lr){3-5}\cmidrule(lr){6-9}
    Dataset & Regime & Match & Mism.(avg) & Mism.(best)
            & Match & Mism.(avg) & Mism.(best) & ZS \\
    \midrule
    \multirow{2}{*}{\shortstack[l]
    {Food-101}}
      & Imbalanced  & $65.7$ & $71.7${\scriptsize$\,(+6.0)$} & $75.7${\scriptsize$\,(+10.0)$} & $81.9$ & $84.8$ & $86.8$ & \multirow{2}{*}{$86.4$} \\
      & Oversampled & $69.9$ & $75.5${\scriptsize$\,(+5.6)$} & $78.9${\scriptsize$\,(+9.0)$}  & $83.9$ & $86.5$ & $88.1$ & \\
    \midrule
    \multirow{2}{*}{\shortstack[l]
    {CUB-200}}
      & Imbalanced  & $40.8$ & $46.3${\scriptsize$\,(+5.5)$} & $59.5${\scriptsize$\,(+18.7)$} & $66.2$ & $65.6$ & $72.3$ & \multirow{2}{*}{$59.7$} \\
      & Oversampled & $57.4$ & $58.3${\scriptsize$\,(+0.9)$} & $71.0${\scriptsize$\,(+13.6)$} & $74.1$ & $70.1$ & $76.9$ & \\
    \midrule
    \multirow{2}{*}{\shortstack[l]
    {FGVC-Aircraft}}
      & Imbalanced  & $35.5$ & $41.0${\scriptsize$\,(+5.5)$} & $46.1${\scriptsize$\,(+10.6)$} & $64.8$ & $65.7$ & $67.4$ & \multirow{2}{*}{$60.8$} \\
      & Oversampled & $41.7$ & $48.8${\scriptsize$\,(+7.1)$} & $56.2${\scriptsize$\,(+14.5)$} & $67.1$ & $67.6$ & $70.1$ & \\
    \midrule
    \multirow{2}{*}{\shortstack[l]
    {Stanford Dogs}}
      & Imbalanced               & $57.0$ & $59.5${\scriptsize$\,(+2.5)$} & $63.1${\scriptsize$\,(+6.1)$} & $76.4$ & $75.3$ & $77.0$ & \multirow{2}{*}{$66.5$} \\
      & Oversampled  & $60.0$ & $61.4${\scriptsize$\,(+1.4)$} & $65.8${\scriptsize$\,(+5.8)$} & $77.7$ & $75.7$ & $78.2$ & \\
\midrule
    \multirow{2}{*}{\shortstack[l]
{ImageNet}} %
& Imbalanced & $45.3$ & $47.9${\scriptsize$\,(+2.6)$} & $53.9${\scriptsize$\,(+8.7)$} & $68.6$ & $69.7$ & $72.6$ & \multirow{2}{*}{$66.6$} \\
& Oversampled & $48.5$ & $54.4${\scriptsize$\,(+5.8)$} & $59.4${\scriptsize$\,(+10.9)$} & $70.2$ & $72.2$ & $74.3$ & \\
\midrule
    \multirow{2}{*}{\shortstack[l]
{PlantNet-300K}} %
 & Imbalanced & $9.8$ & $13.6${\scriptsize$\,(+3.8)$} & $19.2${\scriptsize$\,(+9.4)$} & $52.9$ & $51.1$ & $54.4$ & \multirow{2}{*}{$47.6$}\\
 & Oversampled & $19.2$ & $28.7${\scriptsize$\,(+9.5)$} & $42.1${\scriptsize$\,(+22.9)$} & $57.2$ & $55.2$ & $59.1$ & \\
\bottomrule
  \end{tabular}
\end{table*}

\noindent\textbf{Rare-class mitigation.}
Figure~\ref{fig:rescue} reports the change in recall from switching to a mismatched prompt, relative to the matched prompt, shown separately for rare and frequent classes on Food-101, CUB-200, Stanford Dogs and FGVC-Aircraft across six models.
We use the same rare/frequent
class assignment as in Section~\ref{sec:suppression}.
Guided by the maximum number of images available per class ($n_f$ is capped by each dataset, ranging from $29$ for CUB to $750$ on Food-101), we set the rare count $n_r$ to $5$ for Food-101, $2$ for Stanford Dogs, $2$ for FGVC-Aircraft, and $1$ for CUB-200 (Section~\ref{sec:dataset_analysis} provides analysis on the effect of different imbalance ratios as $n_r$ varies).
The figure supports two conclusions.
First, the averaged mismatch gain on rare recall is positive in all $24$ dataset $\times$ model cells, rising further under the best mismatch.
Second, the effect is specific to the rare classes, and comes with a frequent-class cost that varies by dataset.

\noindent\textbf{Oversampling.}
A natural mitigation for class imbalance is to oversample the rare classes.
We therefore investigate how cross-prompt inference interacts with oversampling.
We compare against random oversampling, which resamples each rare class up to the frequent count without adding new images.
Table~\ref{tab:macrofull} reports rare and overall recall under the matched prompt, the averaged mismatch, and the best mismatch, for both the standard imbalanced finetuning and the oversampled one.

Notably, cross-prompt inference and oversampling are complementary. On the oversampled model,
the averaged mismatch gain in rare recall ranges from $+0.9$ to $+9.5$~pp across datasets,
and the best mismatch gain from $+5.8$ to $+22.9$~pp.
Overall recall (measured over all classes) improves under the best mismatch on every dataset, while the averaged mismatch trades some overall recall on CUB-200, Stanford Dogs, and PlantNet-300K, in both regimes.
Averaged cross-prompt inference also exceeds the zero-shot (pretrained) model in overall recall on every dataset.

Appendix~\ref{app:lora} repeats the analysis with LoRA finetuning, showing that the
matched-prompt penalty also persists under less aggressive adaptation and that
cross-prompt inference mitigates it there too.

\begin{figure}
  \centering
  \begin{subfigure}[t]{0.46\linewidth}
    \centering
    \includegraphics[width=\linewidth]{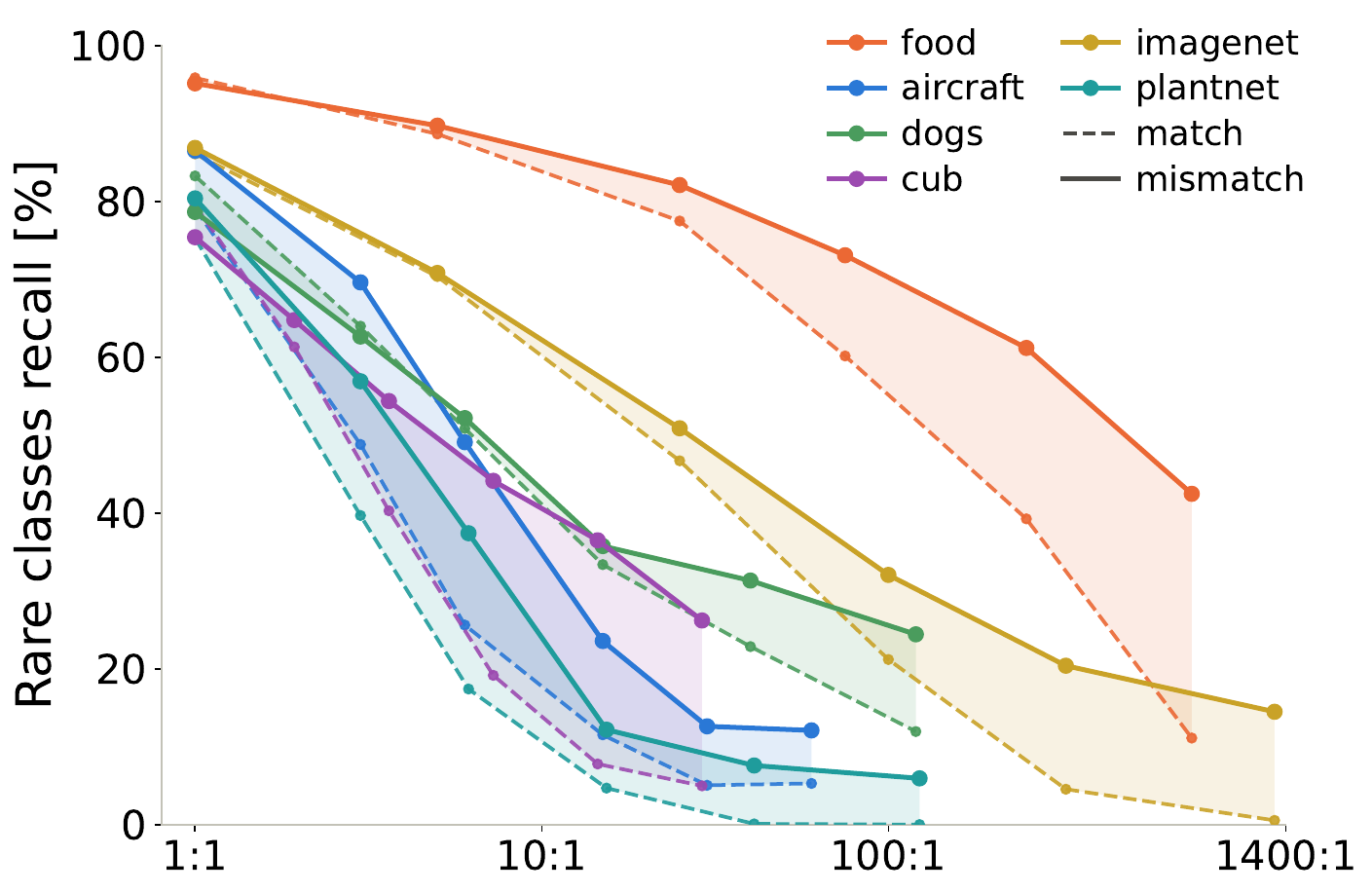}
    \caption{Rare class accuracy vs.\ imbalance ratio (F:R).}
    \label{fig:rare_recall_ratio}
  \end{subfigure}
  \begin{subfigure}[t]{0.46\linewidth}
    \centering
    \includegraphics[width=\linewidth]{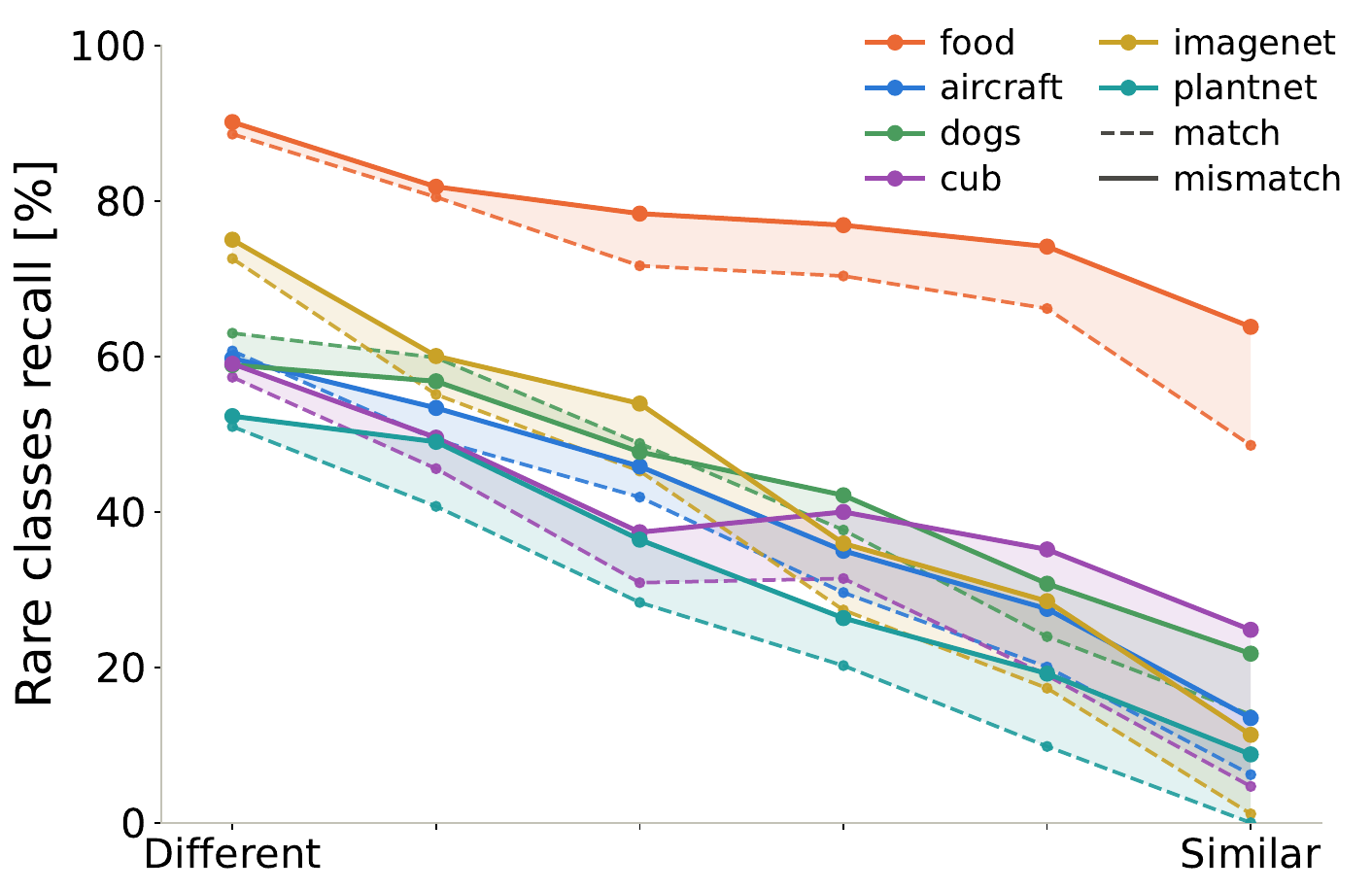}
    \caption{Rare class accuracy vs.\ F-R similarity level.}
    \label{fig:rare_recall_similarity}
  \end{subfigure}
  \caption{%
    \textbf{Rare-class degradation:}
    Dashed lines show the average accuracy over matched prompts.
    Solid lines show the average accuracy over best mismatch prompts.
  }
  \label{fig:rare_recall_two_axes}
\end{figure}

\subsection{Rare-Class Degradation: Imbalance and Class Similarity}
\label{sec:dataset_analysis}
In this section we characterize rare-class degradation, and its recovery using cross-prompt inference, as a function of two factors: (1) the frequent-to-rare imbalance ratio, and (2) the visual similarity between the frequent and rare classes.
For each of the six datasets, we use SigLIP2~\citep{siglip2} to compute pairwise inter-class similarity and select ten pairs of similar classes, yielding 20 classes per dataset.
In order to analyze the similarity factor, we fix the imbalance ratio and selected different frequent-rare class splits in order to control the visual similarity level between rare and frequent classes.
Given the 20 selected classes per dataset, we construct six frequent-rare class splits with progressively decreasing visual similarity between the frequent and rare groups. The first split pairs the most visually similar classes across the two groups, while subsequent splits progressively reduce the inter-group similarity, as measured by SigLIP2~\citep{siglip2}.
In the ratio imbalance analysis experiment, we fix the frequent-rare partition and gradually remove rare samples.
Full implementation details are given in~\ref{app:rare_factor_imp}.

Figure~\ref{fig:rare_recall_ratio} shows the rare-class recall declining monotonically as the imbalance ratio grows, on all six datasets.
We can observe that the rate at which rare-class recall degrades varies between datasets.
CUB-200 and FGVC-Aircraft rare-recall has collapsed by roughly $30{:}1$, whereas Food-101 retains most of its rare-recall even beyond $100{:}1$.
When varying the similarity between rare and frequent classes, while leaving the rare-to-frequent ratio fixed, we can also measure an almost monotonic degradation in rare-class recall in~\ref{fig:rare_recall_similarity}.

Across both experiments, the gap between matched-prompt and cross-prompt inference closely tracks the degradation in rare-class recall.
When rare-class recall remains high, at low imbalance ratios or under dissimilar partitions, the two settings largely coincide.
As recall degrades, however, cross-prompt inference increasingly recovers the lost performance, reaching gains of $+31.4$~pp on Food-101 at $750{:}1$ and $+20.2$~pp on CUB-200 at its most similar partition.
These results suggest that cross-prompt benefits grow with rare-class suppression induced by imbalanced fine-tuning, which is jointly governed by the imbalance ratio and the similarity between frequent and rare classes.

\begin{figure}[t]
    \centering
    \begin{minipage}[t]{0.64\linewidth}
        \vspace{0pt}
        \centering
        \begin{subfigure}[b]{0.49\linewidth}
            \includegraphics[width=\linewidth]{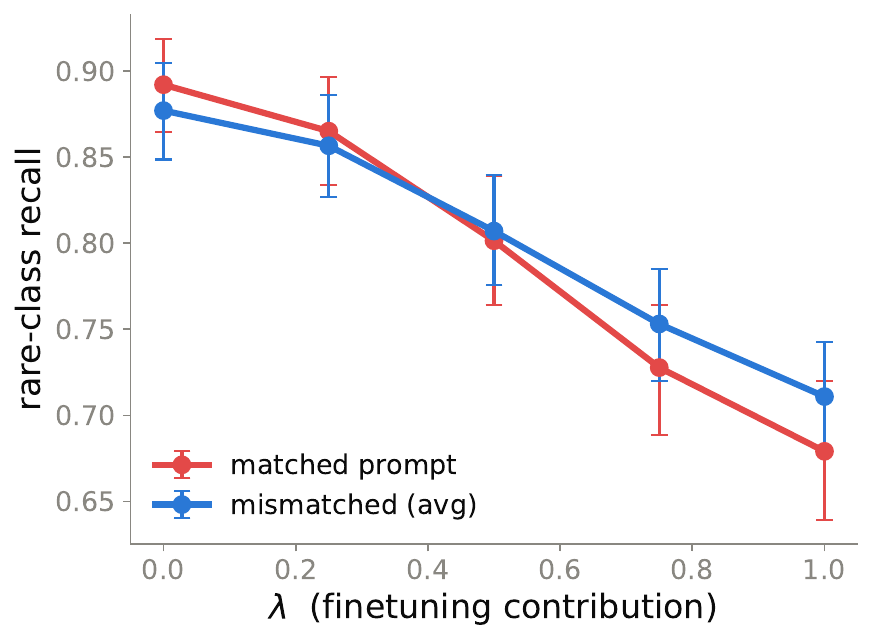}
            \caption{Food-101}
        \end{subfigure}
        \hfill
        \begin{subfigure}[b]{0.49\linewidth}
            \includegraphics[width=\linewidth]{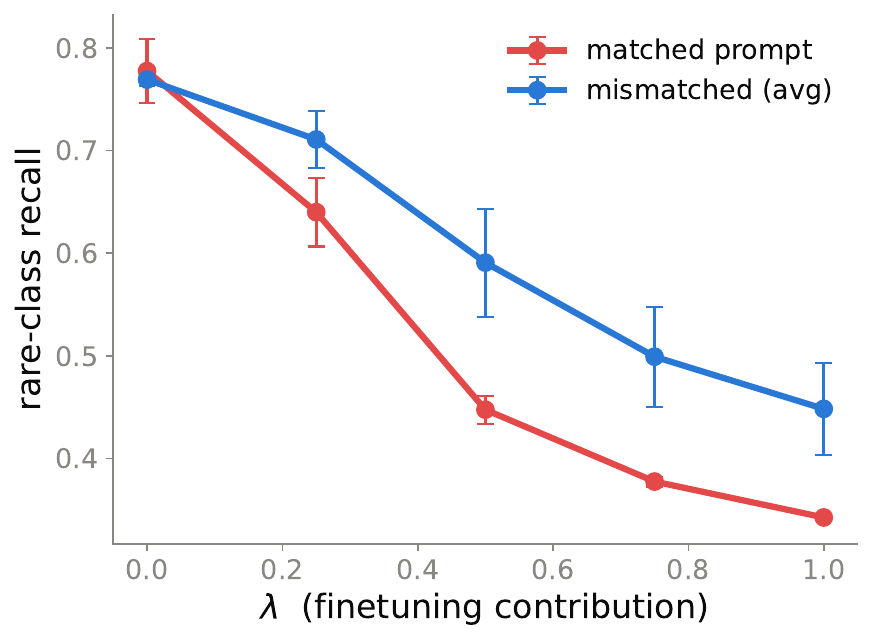}
            \caption{FGVC-Aircraft}
        \end{subfigure}
    \end{minipage}
    \hfill
    \begin{minipage}[t]{0.33\linewidth}
        \vspace{0pt}
        \vspace{-\abovecaptionskip}
        \caption{\textbf{Finetuning contribution and cross-prompt recovery.}
        Rare-class recall as $\lambda$ rescales the finetuning component. The mismatch
        advantage is absent once the finetuning component is removed.}
        \label{fig:tuco-lam}
    \end{minipage}
    \vspace{-0.5cm}
\end{figure}

\subsection{TuCo Analysis Results}
\label{sec:tuco_results}

In this section, we test the predictions of Section~\ref{sec:tuco}. For each dataset we
finetune four models with different descriptive prompts and evaluate each under
its training prompt (matched) and the remaining six prompts (mismatched). 
We measure rare-class recall as a function of $\lambda$, and separately the TuCo
score under matched and mismatched evaluation, reported per model in
Appendix~\ref{app:tuco}.

Figure~\ref{fig:tuco-lam} shows rare-class recall against $\lambda$ on Food-101
and FGVC-Aircraft.
First, matched rare-class recall
rises steeply as the finetuning component is removed, from $67.9\%$ to $89.2\%$
on Food-101 and from $34.3\%$ to $77.7\%$ on FGVC-Aircraft, showing that the
computation introduced by imbalanced finetuning is a major source of rare-class
suppression. Second, the advantage of mismatched evaluation is absent at
$\lambda=0$ and appears once the finetuning component is active, giving a
finetuning-attributable advantage of $+4.7$~pp on Food-101 and $+11.4$~pp on
FGVC-Aircraft. On Food-101 the two curves cross at $\lambda \approx 0.4$, so
imbalanced finetuning inverts which prompt is preferable, while on FGVC-Aircraft
they start together and separate immediately. Results on CUB-200 and on Stanford Dogs are in Appendix~\ref{app:tuco}.

We additionally compute the TuCo score itself, directly testing whether
mismatched evaluation engages the finetuning component less
(Appendix~\ref{app:tuco}, Table~\ref{tab:tuco}). Across datasets, TuCo is lower
under mismatch in all sixteen finetuned models and on $86\%$ of test images on
average ($78$--$93\%$ per dataset). The matched prompt yields the highest
finetuning contribution among all seven evaluation prompts in thirteen of the
sixteen models. The reduction is not specific to rare classes, as TuCo decreases
comparably for frequent classes, indicating a global shift in finetuning
influence.

Together, these analyses are consistent with the hypothesis of Section~\ref{sec:tuco} that cross-prompt inference shifts computation away from imbalanced finetuning and toward the pretrained component, while the rare-class advantage is substantially reduced when the finetuning component is removed.
\vspace{-0.1cm}

\subsection{Non-Descriptive Prompts}
\label{sec:non_descriptive}

In this section, we examine whether the cross-prompt gain requires the class descriptions or arises from any prompt mismatch. Using the same controlled setup as before, we repeat the analysis with six non-descriptive prompts that list the same class names but add no per-class descriptions, differing only in phrasing and format, such as a plain question, a numbered list, or an expert persona. Table~\ref{tab:neutral} compares the cross-prompt gain under descriptive and non-descriptive prompts. Switching between two non-descriptive prompts does not improve rare recall. 

This suggests that the penalty requires descriptive content in the prompts, and that cross-prompt inference can mitigate it, since two non-descriptive prompts differ only in phrasing.

In Appendix~\ref{app:blocks} we further split the grid by whether descriptions are used at finetuning, at evaluation, or both. The matched descriptive pairing is the weakest descriptive setting, and switching to a different descriptive prompt improves rare recall, suggesting that the mismatch matters most, not which side carries the descriptions.

\begin{table}[t]
  \centering
\caption{\textbf{Non-descriptive prompt setting.} Rare-class recall (\%), averaged over prompts, for matched vs.\ mismatched evaluation, with the cross-prompt gain $\Delta$.
}
  \label{tab:neutral}
  \small
  \setlength{\tabcolsep}{5pt}
  \begin{tabular}{@{}l ccc ccc@{}}
    \toprule
     & \multicolumn{3}{c}{Descriptive} & \multicolumn{3}{c}{Non-descriptive} \\
    \cmidrule(lr){2-4}\cmidrule(lr){5-7}
    Dataset & Match & Mism.\ & $\Delta$ & Match & Mism.\ & $\Delta$ \\
    \midrule
    Food-101       & $65.7$ & $71.7$ & $+6.0$ & $64.8$ & $60.7$ & $-4.1$ \\
    CUB-200        & $40.8$ & $46.3$ & $+5.5$ & $39.0$ & $37.7$ & $-1.3$ \\
    FGVC-Aircraft  & $35.5$ & $41.0$ & $+5.5$ & $35.4$ & $33.5$ & $-1.9$ \\
    Stanford Dogs  & $57.0$ & $59.5$ & $+2.5$ & $55.7$ & $55.9$ & $+0.2$ \\

    \bottomrule
  \end{tabular}
\end{table}

\subsection{Leveraging Cross-Prompt Inference}
\label{sec:ensemble}
In this section, we show that our observations on the matched-prompt
penalty can be used to mitigate class imbalance in VLM classification.
Specifically, we average the per-class
scores produced by the same finetuned model under multiple
evaluation prompts and predict the class with the highest
average score. This provides a practical way to apply
cross-prompt inference.

Table~\ref{tab:ens} shows that ensembling improves rare-class
recall on all four datasets relative to matched-prompt inference,
at roughly half the average frequent-class cost of individual mismatched prompts. We evaluate two ensemble
variants, with and without the matched prompt. Including the
matched prompt yields smaller rare-class gains and better
preserves frequent-class recall.

In Appendix~\ref{app:allmethods} we extend this comparison with a variant that selects a
single mismatched prompt using a held-out validation set of one or five labeled images per
rare class. We evaluate all variants with and without oversampling, and find that
combining cross-prompt inference with oversampling achieves the highest overall recall
among the evaluated configurations on all four datasets.

\begin{table}[!t]
  \centering
\caption{\textbf{Prompt ensembling.} Changes in rare and frequent
recall relative to the matched prompt (gain $\Delta$)
for Mism.(avg) and prompt ensembling.}
  \label{tab:ens}
  \small
  \begin{tabular}{@{}lccccccc@{}}
    \toprule
    & & \multicolumn{2}{c}{Mism.(avg)} & \multicolumn{2}{c}{Ens.(mism)} & \multicolumn{2}{c}{Ens.(all)} \\
    \cmidrule(lr){3-4}\cmidrule(lr){5-6}\cmidrule(lr){7-8}
    Dataset & Match & Rare & Freq. & Rare & Freq. & Rare & Freq. \\
    \midrule
    Food-101       & $65.7$ & $+6.0$ & $-0.2$ & $+7.0$ & $-0.3$ & $+6.1$ & $-0.2$ \\
    CUB-200        & $40.8$ & $+5.5$ & $-6.6$ & $+5.5$ & $-3.7$ & $+3.7$ & $-2.9$ \\
    FGVC-Aircraft  & $35.5$ & $+5.5$ & $-3.6$ & $+3.9$ & $-1.3$ & $+2.4$ & $-0.6$ \\
    Stanford Dogs  & $57.0$ & $+2.5$ & $-4.7$ & $+4.6$ & $-2.5$ & $+4.1$ & $-1.8$ \\
    \bottomrule
  \end{tabular}
\end{table}

\section{Limitation}
\label{sec:limitation}
Our experiments use controlled $20$-class subsets so that per-class descriptions remain practical to include in the VLM prompt. Extending cross-prompt inference to settings with hundreds or thousands of classes is an important direction for future work, where class descriptions may exceed practical context lengths. In addition, the effect is specific to the imbalance-forgetting regime. When finetuning does not suppress rare-class performance, cross-prompt inference can provide little benefit or even hurt. This suggests that cross-prompt inference is most valuable when imbalanced finetuning suppresses knowledge the pretrained model already had.

\section{Conclusion}
\label{sec:summary}

We identify the matched-prompt penalty in class-imbalanced finetuning
of VLMs. Although the natural choice is to evaluate
with the prompt used for finetuning, we find that this matched prompt
can be suboptimal for rare classes, while a different descriptive
prompt can recover suppressed rare-class recall without retraining.
We call this inference-time strategy cross-prompt inference.
Its benefit varies with the imbalance ratio and the visual similarity
between frequent and rare classes, and is largest where finetuning
has substantially suppressed rare-class recall.
Our analysis supports the interpretation that cross-prompt inference
draws more on pretrained class knowledge. Specifically, Tuning
Contribution (TuCo) reveals a lower relative contribution from
finetuning-induced computation under mismatched prompts, and removing
this component also removes the cross-prompt advantage.
An extensive study across six datasets and six VLMs supports these findings.

\section*{Reproducibility Statement}
Appendix~\ref{app:impl} provides the full training recipe, the public
checkpoints we finetune, and the log-probability decision rule, and the same
configuration is used for every dataset, prompt, seeds, imbalance ratio and backbone
with no hyperparameter search.
Appendix~\ref{app:lora} specifies the LoRA configuration used for the adaptation-method control.
Table~\ref{tab:datasets} and
Appendix~\ref{app:rare_factor_imp} describe how each imbalanced subset is built,
including the 20 classes per dataset, the SigLIP2 pairing of visually similar
classes, the frequent and rare sample counts, and the imbalance ratio and similarity
sweeps. Appendix~\ref{app:prompts} reproduces all seven prompts and every per-class
description in full, so each train and evaluation matrix can be rebuilt exactly.
Appendix~\ref{app:tuco} specifies the TuCo decomposition and its forward hook
implementation together with the $\lambda$ sweep and the evaluation subsets. The
seeding protocol is described in Section~\ref{sec:suppression}, and all six datasets
and all six model checkpoints are publicly available.

\section*{AI Use Statement}
We used generative AI tools to assist with refining and polishing parts of the paper, figure creation, and software code. 
We also used a large language model to draft the per-class descriptions in
Appendix~\ref{app:prompts} from the class names and dataset taxonomies, and the
authors verified them against the dataset documentation before use.
AI-assisted code was validated and tested by the authors. AI-assisted text and figures were reviewed and edited by the authors. The authors take full responsibility for the final content of the paper.

\bibliography{iclr2027_conference}
\bibliographystyle{iclr2027_conference}

\clearpage
\appendix

\section{Implementation Details}
\label{app:impl}

\paragraph{Training.} All six backbones share one recipe: the Qwen3-VL ladder at
2B, 4B and 8B (\texttt{Qwen/Qwen3-VL-\{2,4,8\}B-Instruct}), Qwen2.5-VL-3B
(\texttt{Qwen/Qwen2.5-VL-3B-Instruct}), Gemma-3-4B (\texttt{google/gemma-3-4b-it})
and InternVL-3.5-2B (\texttt{OpenGVLab/InternVL3\_5-2B-HF}).
Models are loaded in bfloat16 with FlashAttention-2 and gradient checkpointing. We fully finetune
the language model and keep the vision encoder and multimodal projector frozen, so for
Qwen3-VL-2B the $1.72$B trainable parameters exclude the $0.41$B of the vision tower.
AdamW, learning rate $1\times10^{-5}$, weight decay $0$, gradient-norm clipping at $1.0$, cosine
schedule with $\min(50,\lfloor T/20\rfloor)$ warmup steps for $T$ total steps. Batch
size $4$ with gradient accumulation $4$ (effective $16$), $6$ epochs. Reported
numbers are from the final epoch: there is no early stopping and no checkpoint
selection. Evaluation batch size is $32$ for the Qwen models, $16$ for
InternVL-3.5-2B and the processor default for Gemma-3-4B, the only setting that
varies and only to fit memory.
Unless stated otherwise, results are for
Qwen3-VL-2B-Instruct.

\begin{table}[h]
  \centering\small
  \caption{Dataset configurations. $n_f$/$n_r$ are images per frequent/rare class.
    $n_f$ is the largest count the source pool supports.}
  \label{tab:datasets}
  \begin{tabular}{@{}lrrrr@{}}
    \toprule
    Dataset & $n_f$ & $n_r$ & ratio & train / test \\
    \midrule
    Food-101      & $750$ & $5$ & $150{:}1$ & $7550$ / $2000$ \\
    CUB-200       & $29$  & $1$ & $29{:}1$  & $300$ / $220$   \\
    FGVC-Aircraft & $60$  & $2$ & $30{:}1$  & $620$ / $600$   \\
    Stanford Dogs & $100$ & $2$ & $50{:}1$  & $1020$ / $600$  \\
    \bottomrule
  \end{tabular}
\end{table}

\paragraph{Decision rule.} 
Classification never uses free-form generation. For each
image we compare the model's log-probability for each candidate class name and take the
argmax, which makes the prediction deterministic.

\paragraph{Reproducibility.} One configuration is used for every dataset, prompt,
seed, imbalance ratio and backbone, with no hyperparameter search. The prompt bank
is fixed before any run (Appendix~\ref{app:prompts}) and descriptions are authored
once per (dataset, class, style).

\section{Finetuning Contribution Analysis: Details}
\label{app:tuco}

Transformer layers update the residual stream additively, so a finetuned layer splits into a
pretrained and a finetuning part. For layer $\ell$ we evaluate the finetuned layer
$f_\ell^{\mathrm{FT}}$ and its pretrained counterpart $f_\ell^{\mathrm{PT}}$ on the same hidden state
$h_\ell$, taken from the finetuned trajectory,
\begin{equation}
  \mathrm{PTC}_\ell = f_\ell^{\mathrm{PT}}(h_\ell) - h_\ell,
  \qquad
  \mathrm{FTC}_\ell = f_\ell^{\mathrm{FT}}(h_\ell) - f_\ell^{\mathrm{PT}}(h_\ell),
  \label{eq:app-ptc-ftc}
\end{equation}
so that $h_{\ell+1} = h_\ell + \mathrm{PTC}_\ell + \mathrm{FTC}_\ell$ reproduces the finetuned
computation exactly. Following \citet{nuti2025tuco}, the tuning contribution at the final prompt
token is
\begin{equation}
  \mathrm{TuCo} =
  \frac{\lVert \sum_\ell \mathrm{FTC}_\ell \rVert_1}
       {\lVert \sum_\ell \mathrm{PTC}_\ell \rVert_1 + \lVert \sum_\ell \mathrm{FTC}_\ell \rVert_1},
  \label{eq:app-tuco}
\end{equation}
with the pretrained contribution given by $1-\mathrm{TuCo}$. Scaling $\mathrm{FTC}_\ell$ by $\lambda$
as in Equation~\ref{eq:tuco_lambda} interpolates the two layers,
$h_{\ell+1} = (1-\lambda) f_\ell^{\mathrm{PT}}(h_\ell) + \lambda f_\ell^{\mathrm{FT}}(h_\ell)$.
\citet{nuti2025tuco} sweep this factor over $[0.75,1.25]$, and we extend it to $\lambda=0$ to remove
the finetuning component. This internal measurement is not equivalent to comparing the two
models' outputs, since finetuning can alter internal representations while leaving outputs unchanged,
and an output-level metric can trend opposite to TuCo~\citep{nuti2025tuco}.

\paragraph{Implementation and setup.}
We implement Equations~\ref{eq:app-ptc-ftc} and~\ref{eq:app-tuco} with forward hooks on the language
decoder layers, so the pretrained layer receives exactly the hidden state, attention mask and
positional encodings of the finetuned trajectory. Running the two models independently would not give
this decomposition, since their trajectories diverge after the first layer. The intervention is
confined to these layers, so the finetuned embeddings, output norm and language modeling head remain
in place, as does the frozen vision tower. 
All runs use Qwen3-VL-2B-Instruct with the protocol of the main experiments (full
finetuning of the language model, six epochs). Per dataset, we finetune four models, one per descriptive prompt, using \texttt{detailed\_all} with the appearance, contrastive and attribute
styles and \texttt{detailed\_rare} with the appearance style, and evaluate each under all seven
prompts. We use $500$ balanced test images ($25$ per class) on Food-101 and $200$ ($10$ per class) on
the other three, sweeping $\lambda \in \{0, 0.25, 0.5, 0.75, 1\}$.

\paragraph{Per-model TuCo.}
Table~\ref{tab:tuco} reports the finetuning contribution under matched and mismatched
prompts for each of the sixteen finetuned models. ``Drop'' is the fraction of test
images on which TuCo decreases under mismatch, and ``rank'' the position of the
matched prompt among all seven evaluation prompts ordered by decreasing TuCo. TuCo is
lower under mismatch in all sixteen models, and the matched prompt has the highest
TuCo in thirteen of them.

\begin{table}[!t]
  \centering
  \caption{\textbf{Finetuning contribution under matched and mismatched prompts.} Mean TuCo for each
  finetuned model under its own training prompt (matched) and averaged over the six remaining prompts
  (mismatched).}
  \label{tab:tuco}
  \small
  \begin{tabular}{@{}lllccc@{}}
    \toprule
    Dataset & Training prompt & Matched & Mismatched & $\Delta$ ($\times10^{-3}$) & Drop / Rank \\
    \midrule
    \multirow{4}{*}{Food-101}
      & detailed\_all\_\_contrastive   & $0.0529$ & $0.0518$ & $-1.07$ & $89.6\%$ / 2 \\
      & detailed\_all\_\_appearance    & $0.0525$ & $0.0515$ & $-0.95$ & $90.4\%$ / 1 \\
      & detailed\_all\_\_attribute     & $0.0547$ & $0.0531$ & $-1.61$ & $99.4\%$ / 1 \\
      & detailed\_rare\_\_appearance   & $0.0523$ & $0.0513$ & $-1.01$ & $93.4\%$ / 1 \\
    \cmidrule(l){1-6}
    \multirow{4}{*}{CUB-200}
      & detailed\_all\_\_contrastive   & $0.0254$ & $0.0243$ & $-1.09$ & $91.5\%$ / 1 \\
      & detailed\_all\_\_appearance    & $0.0251$ & $0.0241$ & $-1.05$ & $92.0\%$ / 1 \\
      & detailed\_all\_\_attribute     & $0.0265$ & $0.0253$ & $-1.18$ & $95.5\%$ / 1 \\
      & detailed\_rare\_\_appearance   & $0.0248$ & $0.0243$ & $-0.46$ & $76.5\%$ / 1 \\
    \cmidrule(l){1-6}
    \multirow{4}{*}{Stanford Dogs}
      & detailed\_all\_\_contrastive   & $0.0251$ & $0.0240$ & $-1.00$ & $91.5\%$ / 1 \\
      & detailed\_all\_\_appearance    & $0.0254$ & $0.0247$ & $-0.62$ & $78.5\%$ / 2 \\
      & detailed\_all\_\_attribute     & $0.0265$ & $0.0259$ & $-1.15$ & $90.0\%$ / 1 \\
      & detailed\_rare\_\_appearance   & $0.0250$ & $0.0244$ & $-0.61$ & $77.0\%$ / 1 \\
    \cmidrule(l){1-6}
    \multirow{4}{*}{FGVC-Aircraft}
      & detailed\_all\_\_contrastive   & $0.0193$ & $0.0187$ & $-0.59$ & $82.5\%$ / 1 \\
      & detailed\_all\_\_appearance    & $0.0195$ & $0.0188$ & $-0.67$ & $90.5\%$ / 1 \\
      & detailed\_all\_\_attribute     & $0.0201$ & $0.0198$ & $-0.27$ & $66.5\%$ / 3 \\
      & detailed\_rare\_\_appearance   & $0.0204$ & $0.0200$ & $-0.34$ & $73.0\%$ / 1 \\
    \bottomrule
  \end{tabular}
\end{table}

\begin{figure}[!t]
  \centering
  \begin{subfigure}[b]{0.45\linewidth}
    \centering
    \includegraphics[width=\linewidth]{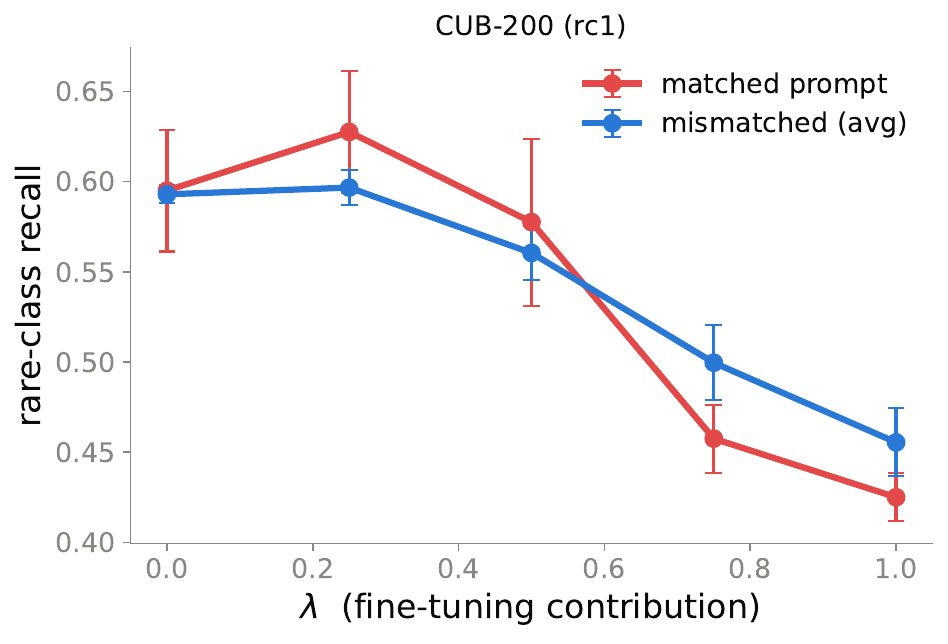}
    \caption{CUB-200}
  \end{subfigure}
  \hfill
  \begin{subfigure}[b]{0.45\linewidth}
    \centering
    \includegraphics[width=\linewidth]{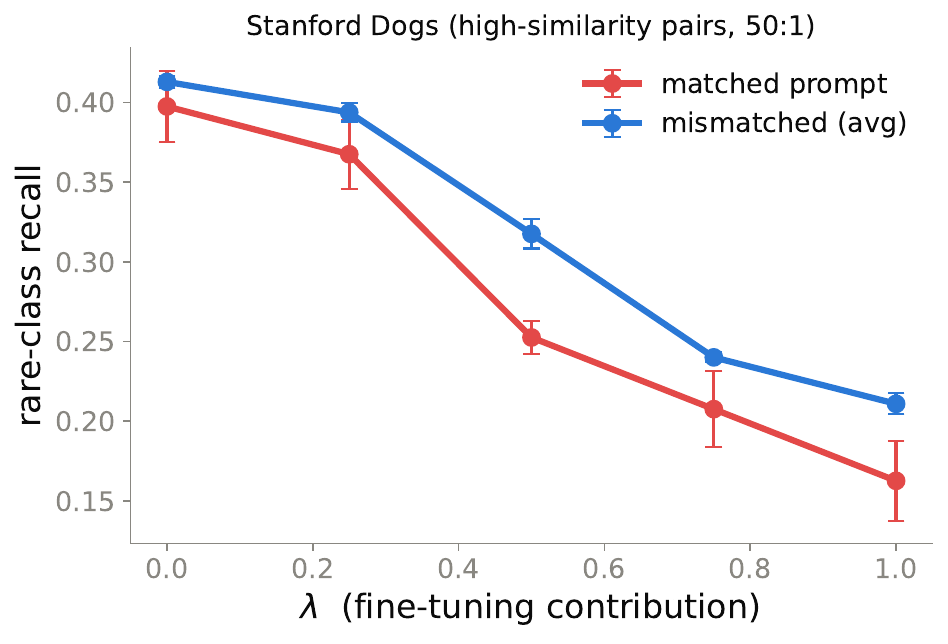}
    \caption{Stanford Dogs}
  \end{subfigure}
\caption{\textbf{Finetuning contribution and cross-prompt recovery on CUB-200 and Stanford Dogs.} Rare-class recall as $\lambda$ rescales the finetuning component. Error bars are standard errors across the four finetuned models.}
  \label{fig:tuco-lam-app}
\end{figure}

\section{Recall Breakdown Across All Models}
\label{app:macrofull}

Table~\ref{tab:macrofull} (Section~\ref{sec:main-result}) summarizes the breakdown on
Qwen3-VL-2B. Table~\ref{tab:macrofull-full} gives a full breakdown across all models. Overall recall is the mean per-class recall over all 20 classes, and rare and frequent recall are the same average restricted to the rare or the frequent classes.
Each Matched and Mismatch cell is averaged over the six descriptive training prompts.

\begin{table*}[t]
  \centering
  \caption{Breakdown across all six models. Each cell reports
    rare\,/\,frequent\,$\cdot$\,overall recall (\%), and the $\Delta$ columns are relative to
    matched.}
  \label{tab:macrofull-full}
  \setlength{\tabcolsep}{5pt}
  \begin{tabular}{@{}llcccc@{}}
    \toprule
    Dataset & Model & Matched & Mismatch & $\Delta$\,r/ovr & $\Delta$best\,r \\
            &       & {\scriptsize R/F$\cdot$Ovr} & {\scriptsize R/F$\cdot$Ovr} & {\scriptsize rare/ovr} & {\scriptsize rare} \\
    \midrule
    Food-101
      & Qwen3-VL-2B     & $65.7/98.2\cdot81.9$ & $71.7/98.0\cdot84.8$ & $+6.0/+2.9$ & $+10.0$ \\
      & Qwen3-VL-4B     & $65.4/98.4\cdot81.9$ & $70.2/98.2\cdot84.2$ & $+4.8/+2.3$ & $+8.4$  \\
      & Qwen3-VL-8B     & $61.0/97.5\cdot79.2$ & $70.1/97.3\cdot83.7$ & $+9.1/+4.5$ & $+13.6$ \\
      & Qwen2.5-VL-3B   & $53.6/96.3\cdot74.9$ & $62.0/95.7\cdot78.9$ & $+8.5/+3.9$ & $+12.1$ \\
      & Gemma-3-4B      & $58.9/97.6\cdot78.2$ & $60.2/97.5\cdot78.9$ & $+1.3/+0.6$ & $+2.9$  \\
      & InternVL-3.5-2B & $38.1/96.0\cdot67.0$ & $47.2/94.4\cdot70.8$ & $+9.1/+3.8$ & $+16.0$ \\
    \midrule
    CUB-200
      & Qwen3-VL-2B     & $40.8/91.5\cdot66.2$ & $46.3/84.9\cdot65.6$ & $+5.5/-0.6$ & $+18.7$ \\
      & Qwen3-VL-4B     & $49.5/90.8\cdot70.2$ & $53.9/87.4\cdot70.7$ & $+4.4/+0.5$ & $+12.3$ \\
      & Qwen3-VL-8B     & $54.4/90.8\cdot72.6$ & $58.4/89.2\cdot73.8$ & $+4.0/+1.2$ & $+7.3$  \\
      & Qwen2.5-VL-3B   & $50.5/88.5\cdot69.5$ & $55.1/84.9\cdot70.0$ & $+4.7/+0.5$ & $+7.9$  \\
      & Gemma-3-4B      & $40.4/88.4\cdot64.4$ & $42.4/85.5\cdot64.0$ & $+2.1/-0.4$ & $+5.6$  \\
      & InternVL-3.5-2B & $29.3/62.7\cdot46.0$ & $38.1/55.4\cdot46.7$ & $+8.8/+0.8$ & $+14.5$ \\
    \midrule
    FGVC-Aircraft
      & Qwen3-VL-2B     & $35.5/94.1\cdot64.8$ & $41.0/90.5\cdot65.7$ & $+5.5/+0.9$ & $+10.6$ \\
      & Qwen3-VL-4B     & $39.3/95.3\cdot67.3$ & $42.0/94.9\cdot68.5$ & $+2.8/+1.2$ & $+5.0$  \\
      & Qwen3-VL-8B     & $35.5/93.4\cdot64.5$ & $38.8/92.2\cdot65.5$ & $+3.2/+1.0$ & $+10.6$ \\
      & Qwen2.5-VL-3B   & $34.4/92.4\cdot63.4$ & $35.9/91.5\cdot63.7$ & $+1.6/+0.3$ & $+3.6$  \\
      & Gemma-3-4B      & $31.5/93.5\cdot62.5$ & $33.1/92.8\cdot62.9$ & $+1.6/+0.4$ & $+2.7$  \\
      & InternVL-3.5-2B & $14.2/46.6\cdot30.4$ & $16.5/42.0\cdot29.2$ & $+2.2/-1.2$ & $+3.8$  \\
    \midrule
    Stanford Dogs
      & Qwen3-VL-2B     & $57.0/95.7\cdot76.4$ & $59.5/91.0\cdot75.3$ & $+2.5/-1.1$ & $+6.1$ \\
      & Qwen3-VL-4B     & $63.9/96.3\cdot80.1$ & $65.7/94.5\cdot80.1$ & $+1.8/+0.0$ & $+4.9$ \\
      & Qwen3-VL-8B     & $64.8/95.9\cdot80.3$ & $65.1/94.7\cdot79.9$ & $+0.3/-0.4$ & $+3.8$ \\
      & Qwen2.5-VL-3B   & $60.8/95.4\cdot78.1$ & $63.0/93.5\cdot78.2$ & $+2.2/+0.1$ & $+5.4$ \\
      & Gemma-3-4B      & $45.3/95.4\cdot70.4$ & $47.8/93.6\cdot70.7$ & $+2.5/+0.3$ & $+4.6$ \\
      & InternVL-3.5-2B & $21.6/74.7\cdot48.2$ & $24.6/66.5\cdot45.6$ & $+2.9/-2.6$ & $+7.0$ \\
    \bottomrule
  \end{tabular}
\end{table*}

\section{Rare-class Suppression}
\label{app:suppression}

\begin{figure}[!t]
    \centering
    \begin{minipage}[t]{0.55\linewidth}
        \vspace{0pt}
        \centering
        \begin{subfigure}[b]{0.49\linewidth}
            \includegraphics[width=\linewidth]{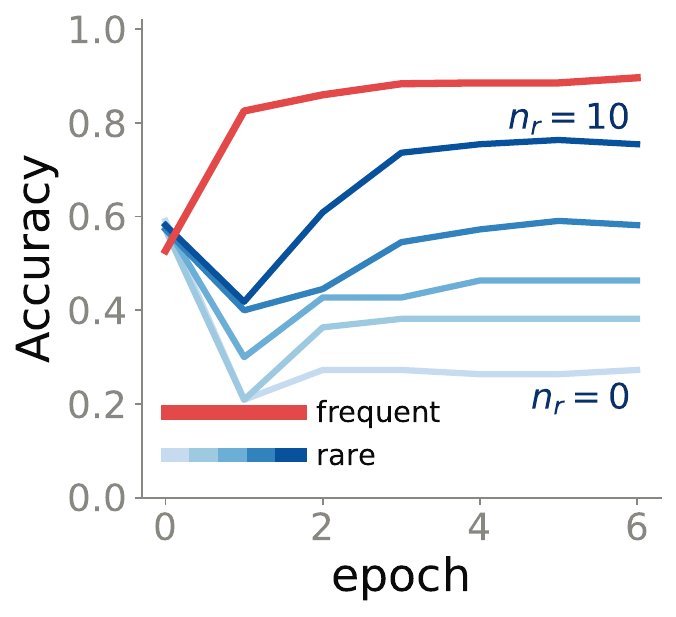}
            \vspace{-3mm}
            \caption{CUB-200}
            \label{fig:forget-cub}
        \end{subfigure}
        \hfill
        \begin{subfigure}[b]{0.49\linewidth}
            \includegraphics[width=\linewidth]{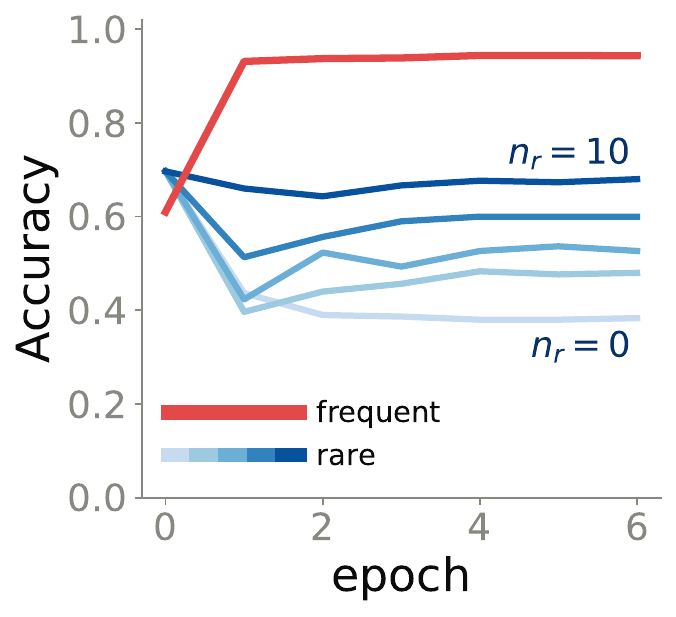}
            \vspace{-3mm}
            \caption{Stanford Dogs}
            \label{fig:forget-dogs}
        \end{subfigure}
    \end{minipage}
    \hfill
    \begin{minipage}[t]{0.44\linewidth}
        \vspace{0pt}
        \vspace{-\abovecaptionskip}
        \caption{\textbf{Rare-class suppression on CUB-200 and Stanford Dogs.}
        Accuracy over finetuning epochs for frequent classes (red) and rare classes
        trained with $n_r \in \{0,1,2,5,10\}$ samples per class (blue, light to dark),
        where $n_r\!=\!0$ trains on frequent classes only. Epoch $0$ is the zero-shot
        model. As in Figure~\ref{fig:rare_suppression}, frequent accuracy saturates
        near ceiling while rare accuracy drops below its zero-shot value, more so as
        $n_r$ decreases.}
        \label{fig:rare_suppression_app}
    \end{minipage}
\end{figure}

Figure~\ref{fig:rare_suppression_app} repeats the suppression experiment of
Section~\ref{sec:suppression} on CUB-200 and Stanford-Dogs, with
an identical setup and the same frequent/rare class assignment. The pattern holds on
both, and the drop deepens as $n_r$ decreases.

\section{Selection Methods for Cross-prompt Inference}
\label{app:allmethods}
Table~\ref{tab:allmethods} presents practical methods that can be used for cross-prompt
inference, with and without oversampling. We use a simple mechanism that picks the
evaluation prompt for each training prompt from a minimal held-out validation set. For
each of $200$ random splits we use $k$ labeled images per rare class for selection and
evaluate recall on the remaining images, then average the results across splits.

Mism.(avg) reports average recall across individual mismatched prompts, Mism.(ens)
averages their per-class scores before prediction, and Mism.(sel, $k$) selects a single
mismatched prompt using this validation protocol.

Cross-prompt inference is complementary to oversampling, and selecting a prompt from as
few as five labeled images per rare class gives the highest overall recall. Ensembling
reaches comparable overall recall but needs multiple forward passes (one per evaluation prompt).

\begin{table*}[t]
  \centering
  \caption{\textbf{Method comparison.} In each cell we report rare / frequent / overall recall
  (\%). The best overall recall per dataset is in bold. Oversampling resamples each rare
  class up to the frequent count without adding new images.}
  \label{tab:allmethods}
  \small
  \setlength{\tabcolsep}{4pt}
  \begin{tabular}{@{}lcccc@{}}
    \toprule
    Method & Food-101 & CUB-200 & FGVC-Aircraft & Stanford Dogs \\
    \midrule
    Matched prompt               & $65.7$ / $98.2$ / $81.9$ & $40.8$ / $91.5$ / $66.2$ & $35.5$ / $94.1$ / $64.8$ & $57.0$ / $95.7$ / $76.4$ \\
    Mism.(avg)                   & $71.7$ / $98.0$ / $84.8$ & $46.3$ / $84.9$ / $65.6$ & $41.0$ / $90.5$ / $65.7$ & $59.5$ / $91.0$ / $75.3$ \\
    Mism.(ens)                   & $72.7$ / $97.9$ / $85.3$ & $46.3$ / $87.8$ / $67.1$ & $39.4$ / $92.8$ / $66.1$ & $61.6$ / $93.2$ / $77.4$ \\
    Mism.(sel, $k{=}1$)          & $73.8$ / $97.8$ / $85.8$ & $56.4$ / $84.1$ / $70.3$ & $42.9$ / $88.8$ / $65.8$ & $61.0$ / $89.9$ / $75.4$ \\
    Mism.(sel, $k{=}5$)          & $74.8$ / $97.9$ / $86.3$ & $58.6$ / $83.5$ / $71.0$ & $45.3$ / $88.8$ / $67.1$ & $62.4$ / $89.5$ / $75.9$ \\
    \midrule
    \multicolumn{5}{@{}l}{\textit{with oversampling}} \\
    Matched prompt               & $69.9$ / $98.0$ / $83.9$ & $57.4$ / $90.7$ / $74.1$ & $41.7$ / $92.5$ / $67.1$ & $60.0$ / $95.5$ / $77.7$ \\
    Mism.(avg)                   & $75.5$ / $97.5$ / $86.5$ & $58.3$ / $81.9$ / $70.1$ & $48.8$ / $86.4$ / $67.6$ & $61.4$ / $90.0$ / $75.7$ \\
    Mism.(ens)                   & $75.6$ / $97.8$ / $86.7$ & $68.2$ / $84.4$ / $76.3$ & $47.4$ / $89.0$ / $68.2$ & $63.3$ / $92.3$ / \textbf{77.8} \\
    Mism.(sel, $k{=}1$)          & $77.1$ / $97.4$ / $87.2$ & $70.4$ / $80.7$ / $75.5$ & $53.3$ / $81.8$ / $67.6$ & $64.1$ / $88.7$ / $76.4$ \\
    Mism.(sel, $k{=}5$)          & $77.8$ / $97.4$ / \textbf{87.6} & $74.0$ / $80.5$ / \textbf{77.2} & $55.4$ / $81.8$ / \textbf{68.6} & $66.3$ / $88.7$ / $77.5$ \\
    \bottomrule
  \end{tabular}
\end{table*}

\section{LoRA Versus Full Finetuning}
\label{app:lora}
In this section, we compare full fine-tuning with LoRA to test whether the matched-prompt penalty is specific to full-parameter adaptation or persists under a less aggressive fine-tuning regime.
We repeat the match and mismatch (cross-prompt inference) experiments with LoRA \citep{hu2022lora}, a low-rank adapter that leaves the base weights untouched and trains far fewer parameters. 
We use rank $16$ with scaling $32$ on the attention and feedforward projections, and keep the data, prompts, epochs and three random seeds identical to the full-finetuning runs, so the two settings differ only in how the model is adapted.

Table~\ref{tab:lora} reports rare-class recall for both adaptation methods. 
The matched-prompt penalty persists under LoRA with comparable magnitude, indicating that it is not an artifact of full-parameter fine-tuning.
On CUB-200, LoRA yields higher matched-prompt recall than full fine-tuning, while the cross-prompt gain remains comparable and is slightly larger.
Overall, these results show that the matched-prompt penalty is not specific to full-parameter finetuning and persists under substantially lighter parameter adaptation.

\begin{table}[t]
  \centering
  \caption{\textbf{LoRA vs. full finetuning.} Rare-class recall (\%) with the mismatch gain ($\Delta$) vs.\
  matched in parentheses, for both full finetuning and LoRA.}
  \label{tab:lora}
  \small
  \setlength{\tabcolsep}{5pt}
  \begin{tabular}{@{}lccc@{}}
    \toprule
    Dataset & Match & Mism.(avg) & Mism.(best) \\
    \midrule
    \multicolumn{4}{@{}l}{\textit{full finetuning}} \\
    Food-101         & $65.7$ & $71.7${\scriptsize$\,(+6.0)$} & $75.7${\scriptsize$\,(+10.0)$} \\
    CUB-200          & $40.8$ & $46.3${\scriptsize$\,(+5.5)$} & $59.5${\scriptsize$\,(+18.7)$} \\
    FGVC-Aircraft    & $35.5$ & $41.0${\scriptsize$\,(+5.5)$} & $46.1${\scriptsize$\,(+10.6)$} \\
    Stanford Dogs    & $57.0$ & $59.5${\scriptsize$\,(+2.5)$} & $63.1${\scriptsize$\,(+6.1)$} \\
    \midrule
    \multicolumn{4}{@{}l}{\textit{LoRA}} \\
    Food-101         & $65.6$ & $69.6${\scriptsize$\,(+3.9)$} & $72.6${\scriptsize$\,(+7.0)$} \\
    CUB-200          & $45.5$ & $51.3${\scriptsize$\,(+5.8)$} & $64.7${\scriptsize$\,(+19.2)$} \\
    FGVC-Aircraft     & $39.3$ & $47.9${\scriptsize$\,(+8.6)$} & $57.6${\scriptsize$\,(+18.3)$} \\
    Stanford Dogs    & $62.7$ & $64.9${\scriptsize$\,(+2.2)$} & $69.4${\scriptsize$\,(+6.7)$} \\
    \bottomrule
  \end{tabular}
\end{table}

\section{Rare-Class Degradation: Imbalance and Class Similarity Implementation Details}
\label{app:rare_factor_imp}
All runs share one backbone and protocol: Qwen3-VL-2B under full finetuning, a fixed bank of seven prompt styles and six training epochs.
We fully finetune the language model and keep the vision encoder and multimodal projector frozen.
We select 20 classes per dataset to be used in the experiments.
For each dataset, we choose ten pairs of classes that are the most visually similar according to SigLIP2~\citep{siglip2}, without using the same class twice.

\textbf{Imbalance ratio between frequent and rare classes.}
We fix all the rare and frequent classes of each dataset by splitting each of the ten pairs, assigning one class to the frequent set and the other to the rare set at random. This yields ten frequent and ten rare classes per dataset, held fixed across all imbalance ratios so that only the number of training images per rare class varies.

\textbf{Similarity between frequent and rare classes.}
First, we fix the imbalance ratio of each dataset (Food-101 $n_r\!=\!5$, FGVC-Aircraft $n_r\!=\!2$, CUB-200 $n_r\!=\!2$, Stanford Dogs $n_r\!=\!1$, ImageNet $n_r\!=\!2$ and PlantNet-300K $n_r\!=\!2$).
Second, we divide each dataset into six groups, each group with a different split between frequent and rare classes, where we control the similarity level between the rare and frequent classes.
Similarity is calculated as the average SigLIP2 embedding cosine similarity of each rare class to its most similar frequent class.
We then permute the frequent-rare split in order to achieve varied similarity levels for different groups using the same classes.
Table~\ref{tab:similarity-scores} shows the similarity scores for each of the six groups across all datasets.

\begin{table}[t]
\centering
\caption{Average rare-to-frequent SigLIP2 cosine similarity for each of the six frequent-rare partitions, ordered from most to least similar. 
All groups use the same 20 classes per dataset.}
\label{tab:similarity-scores}
\begin{tabular}{@{}lrrrrrr@{}}
\toprule
Dataset & Group\#1 & Group\#2 & Group\#3 & Group\#4 & Group\#5 & Group\#6 \\
 & (similar) & & & & & (different) \\
\midrule
Food-101       & 0.733 & 0.718 & 0.705 & 0.694 & 0.679 & 0.667 \\
FGVC-Aircraft  & 0.803 & 0.793 & 0.772 & 0.760 & 0.754 & 0.734 \\
Stanford Dogs  & 0.775 & 0.764 & 0.751 & 0.743 & 0.729 & 0.713 \\
CUB-200        & 0.849 & 0.817 & 0.788 & 0.782 & 0.761 & 0.741 \\
ImageNet       & 0.798 & 0.770 & 0.749 & 0.722 & 0.689 & 0.675 \\
PlantNet-300K  & 0.850 & 0.827 & 0.800 & 0.780 & 0.748 & 0.713 \\
\bottomrule
\end{tabular}
\end{table}

\section{Descriptive Prompts at Finetuning and Evaluation}
\label{app:blocks}

In this section we further split the grid to isolate the effect of using descriptive
prompts during finetuning versus at evaluation. We write P for the plain,
description-free prompt and D for any of the six descriptive prompts.
Table~\ref{tab:blocks} reports rare recall for each of the five resulting
configurations, averaged over prompts and seeds.

Switching to a different descriptive prompt improves rare recall on all four datasets,
so the matched pairing is never the best use of a descriptive prompt. Which side
benefits from the descriptions is dataset dependent. Food-101 and Stanford Dogs do
better with descriptions only during finetuning, while CUB-200 and FGVC-Aircraft do
better with descriptions only at evaluation. The penalty therefore follows from using the same prompt at
both ends rather than from where the descriptions sit.

The last two columns also account for the mismatch average in
Table~\ref{tab:neutral}. That average runs over all six mismatched evaluation prompts,
five of them descriptive and one of them plain, so it falls slightly below the
descriptive-only value reported here.

\begin{table}[t]
  \centering
  \caption{\textbf{Where the descriptions appear.} Rare-class recall (\%) averaged over
  prompts and seeds, with P the plain prompt and D a descriptive one. }
  \label{tab:blocks}
  \small
  \setlength{\tabcolsep}{5pt}
  \begin{tabular}{@{}lccccc@{}}
    \toprule
    Dataset & P$\to$P & P$\to$D & D$\to$P & D$\to$D matched & D$\to$D$'$ mismatched \\
    \midrule
    Food-101       & $65.5$ & $70.8$ & $72.7$ & $65.7$ & $71.5$ \\
    CUB-200        & $37.9$ & $52.1$ & $28.5$ & $40.8$ & $49.8$ \\
    FGVC-Aircraft  & $34.3$ & $42.6$ & $39.8$ & $35.5$ & $41.2$ \\
    Stanford Dogs  & $52.3$ & $55.7$ & $60.2$ & $57.0$ & $59.4$ \\
    \bottomrule
  \end{tabular}
\end{table}

\section{Seed Variability}
\label{app:zs}

Table~\ref{tab:se} reports rare-class recall with the gain over the matched prompt along with the gain's standard error across seeds.

\begin{table}[t]
  \centering
  \caption{\textbf{Rare-class recall (\%) with seed variability.} Values in brackets are
  the gain over the matched prompt and the gain's standard error across seeds.}
  \label{tab:se}
  \small
  \setlength{\tabcolsep}{5pt}
  \begin{tabular}{@{}llccc@{}}
    \toprule
    Dataset & Regime & Match & Mism.(avg) & Mism.(best) \\
    \midrule
    \multirow{2}{*}{Food-101}
      & Imbalanced   & $65.7$ & $71.7$ {\scriptsize$(+6.0 \pm 0.8)$} & $75.7$ {\scriptsize$(+10.0 \pm 1.2)$} \\
      & Oversampled  & $69.9$ & $75.5$ {\scriptsize$(+5.6 \pm 0.3)$} & $78.9$ {\scriptsize$(+9.0 \pm 0.6)$} \\
    \midrule
    \multirow{2}{*}{CUB-200}
      & Imbalanced   & $40.8$ & $46.3$ {\scriptsize$(+5.5 \pm 0.8)$} & $59.5$ {\scriptsize$(+18.7 \pm 0.9)$} \\
      & Oversampled  & $57.4$ & $58.3$ {\scriptsize$(+0.9 \pm 1.1)$} & $71.0$ {\scriptsize$(+13.6 \pm 1.6)$} \\
    \midrule
    \multirow{2}{*}{FGVC-Aircraft}
      & Imbalanced   & $35.5$ & $41.0$ {\scriptsize$(+5.5 \pm 0.7)$} & $46.1$ {\scriptsize$(+10.6 \pm 0.5)$} \\
      & Oversampled  & $41.7$ & $48.8$ {\scriptsize$(+7.1 \pm 0.6)$} & $56.2$ {\scriptsize$(+14.5 \pm 0.6)$} \\
    \midrule
    \multirow{2}{*}{Stanford Dogs}
      & Imbalanced   & $57.0$ & $59.5$ {\scriptsize$(+2.5 \pm 0.6)$} & $63.1$ {\scriptsize$(+6.1 \pm 1.2)$} \\
      & Oversampled  & $60.0$ & $61.4$ {\scriptsize$(+1.4 \pm 0.4)$} & $65.8$ {\scriptsize$(+5.8 \pm 0.9)$} \\
    \bottomrule
  \end{tabular}
\end{table}

\section{Prompt Grid Specification}
\label{app:prompts}

We provide the full prompt specification for every class in every dataset so that the method
can be reproduced. Each dataset uses 20 classes, 10 frequent and 10 rare. We use seven
prompts in total. The plain prompt lists the class names only, and the remaining six are
descriptive prompts formed by three description styles (appearance, contrastive, and
attribute), each applied either to all 20 classes or to the rare classes only.

\paragraph{Food-101.} Classes: \texttt{filet mignon}, \texttt{prime rib}, \texttt{pork chop}, \texttt{steak}, \texttt{bolognese spaghetti}, \texttt{carbonara spaghetti}, \texttt{chocolate cake}, \texttt{mousse}, \texttt{bread pudding}, \texttt{french toast}, \texttt{cheesecake}, \texttt{panna cotta}, \texttt{ravioli}, \texttt{dumplings}, \texttt{pizza}, \texttt{sushi}, \texttt{hamburger}, \texttt{ice cream}, \texttt{caesar salad}, \texttt{guacamole}. 
Descriptions are given in Table~\ref{tab:desc-food}.

{\footnotesize
\setlength{\tabcolsep}{4pt}
\renewcommand{\arraystretch}{1.05}
\begin{longtable}{@{}p{0.18\linewidth}p{0.23\linewidth}p{0.30\linewidth}p{0.22\linewidth}@{}}
\caption{\textbf{Food-101: per-class descriptions for all three styles.} A \texttt{detailed\_all} prompt supplies the whole column; \texttt{detailed\_rare} supplies only the rows for rare classes.}\label{tab:desc-food} \\
\toprule
Class & \texttt{appearance} & \texttt{contrastive} & \texttt{attribute} \\
\midrule
\endfirsthead
\multicolumn{4}{@{}l}{\itshape Table~\thetable\ continued from the previous page.}\\
\toprule
Class & \texttt{appearance} & \texttt{contrastive} & \texttt{attribute} \\
\midrule
\endhead
\midrule
\multicolumn{4}{r@{}}{\itshape Continued on the next page.}\\
\endfoot
\bottomrule
\endlastfoot
filet mignon & a thick round cut of grilled beef steak & a tall, round, boneless beef medallion (unlike the flat wide steak or bone-in pork chop) & beef; boneless; shape: tall round medallion; color: dark seared outside, red inside \\
prime rib & a large roasted beef rib cut, pink in the middle & a roasted beef slab with a bone and a pink center (unlike a seared boneless filet) & beef; bone-in; shape: thick roasted slab; color: brown crust, pink center \\
pork chop & a grilled bone-in pork cut & a pale bone-in pork cut, lighter than beef (unlike the dark red steak) & pork; bone-in; shape: flat chop; color: pale tan \\
steak & a grilled slab of beef & a flat, wide, dark-seared beef slab (unlike the tall round filet mignon) & beef; boneless; shape: flat wide slab; color: dark grilled \\
bolognese spaghetti & spaghetti with a red meat sauce & spaghetti under a chunky red tomato-meat sauce (unlike the pale creamy carbonara) & pasta; sauce: red tomato-meat; texture: chunky; color: red \\
carbonara spaghetti & spaghetti with a creamy egg-and-bacon sauce & spaghetti in a pale glossy egg-cream sauce, no red (unlike red bolognese) & pasta; sauce: egg-cream; texture: glossy; color: pale yellow \\
chocolate cake & a dark brown layered chocolate dessert & a firm sliceable brown cake with layers (unlike the soft spooned mousse) & dessert; form: firm sliced layers; base: cake; color: dark brown \\
mousse & a soft whipped chocolate dessert in a cup & a soft airy chocolate cream in a glass, holds no shape (unlike sliceable chocolate cake) & dessert; form: soft airy cream; vessel: glass; color: brown \\
bread pudding & a baked custard-soaked bread dessert & a browned baked bread-custard block (unlike flat pan-fried french toast slices) & dessert; form: baked custard-soaked bread block; color: golden brown \\
french toast & egg-battered fried bread slices & flat pan-fried egg-dipped bread slices, often stacked (unlike the baked bread pudding block) & dessert; form: flat fried bread slices; often stacked; color: golden \\
cheesecake & a pale dense cake with a crumb base & a dense pale wedge on a brown crust base (unlike the crustless molded panna cotta) & dessert; form: dense wedge; base: crumb crust; color: pale \\
panna cotta & a white molded cream dessert & a glossy crustless molded cream dome, often with sauce on top (unlike crusted cheesecake) & dessert; form: molded cream dome; base: none; color: white, glossy \\
ravioli & square stuffed pasta pillows & flat square stuffed pasta parcels in sauce (unlike rounded pinched dumplings) & pasta; form: flat square parcels; filling: yes; color: pale in sauce \\
dumplings & small steamed or boiled filled dough parcels & rounded pinched dough parcels, steamed or boiled (unlike flat square ravioli) & dough; form: rounded pinched parcels; cooking: steamed/boiled; color: pale \\
pizza & a round flatbread with tomato, cheese, and toppings & a round flatbread with tomato, cheese, and toppings & a round flatbread with tomato, cheese, and toppings \\
sushi & rice-and-fish pieces, often with seaweed & rice-and-fish pieces, often with seaweed & rice-and-fish pieces, often with seaweed \\
hamburger & a beef patty in a round bun & a beef patty in a round bun & a beef patty in a round bun \\
ice cream & scoops of frozen dessert & scoops of frozen dessert & scoops of frozen dessert \\
caesar salad & romaine lettuce with croutons and dressing & romaine lettuce with croutons and dressing & romaine lettuce with croutons and dressing \\
guacamole & a green mashed avocado dip & a green mashed avocado dip & a green mashed avocado dip \\
\end{longtable}
}

\paragraph{CUB-200.} Classes: \texttt{california gull}, \texttt{herring gull}, \texttt{ring-billed gull}, \texttt{western gull}, \texttt{arctic tern}, \texttt{caspian tern}, \texttt{common tern}, \texttt{black-throated sparrow}, \texttt{chipping sparrow}, \texttt{field sparrow}, \texttt{song sparrow}, \texttt{baltimore oriole}, \texttt{hooded oriole}, \texttt{orchard oriole}, \texttt{blue jay}, \texttt{cardinal}, \texttt{painted bunting}, \texttt{belted kingfisher}, \texttt{ruby-throated hummingbird}, \texttt{scissor-tailed flycatcher}. Descriptions are given in Table~\ref{tab:desc-cub}.

{\footnotesize
\setlength{\tabcolsep}{4pt}
\renewcommand{\arraystretch}{1.05}
\begin{longtable}{@{}p{0.18\linewidth}p{0.23\linewidth}p{0.30\linewidth}p{0.22\linewidth}@{}}
\caption{\textbf{CUB-200: per-class descriptions for all three styles.} A \texttt{detailed\_all} prompt supplies the whole column; \texttt{detailed\_rare} supplies only the rows for rare classes.}\label{tab:desc-cub} \\
\toprule
Class & \texttt{appearance} & \texttt{contrastive} & \texttt{attribute} \\
\midrule
\endfirsthead
\multicolumn{4}{@{}l}{\itshape Table~\thetable\ continued from the previous page.}\\
\toprule
Class & \texttt{appearance} & \texttt{contrastive} & \texttt{attribute} \\
\midrule
\endhead
\midrule
\multicolumn{4}{r@{}}{\itshape Continued on the next page.}\\
\endfoot
\bottomrule
\endlastfoot
california gull & a medium gull with a yellow bill, dark eye, and grey back & a MEDIUM gull with yellow legs and a dark eye (unlike the pink-legged herring gull) & gull; size: medium; legs: yellow; eye: dark; back: grey \\
herring gull & a large pale-grey gull with pink legs and a red-spotted yellow bill & a LARGE pale gull with PINK legs (unlike the yellow-legged california/ring-billed gulls) & gull; size: large; legs: pink; bill: yellow with red spot; back: pale grey \\
ring-billed gull & a small gull with a yellow bill circled by a black ring, yellow legs & a small gull whose yellow bill has a clean BLACK RING (unlike the red-spotted herring bill) & gull; size: small; bill: yellow with black ring; legs: yellow \\
western gull & a large gull with a very dark slate back and pink legs & a gull with a very DARK slate back (unlike the pale-grey herring and california gulls) & gull; size: large; back: dark slate; legs: pink \\
arctic tern & a slender grey tern with a full black cap and short red legs & a tern with an ALL-RED short bill and very short legs (unlike caspian's thick bill or common's black-tipped bill) & tern; bill: all red; legs: very short red; cap: full black \\
caspian tern & a large tern with a thick bright-red bill and a black cap & a large tern with a THICK bright-red carrot bill (unlike the slender arctic/common terns) & tern; size: large; bill: thick bright red; cap: black \\
common tern & a grey tern with a black-tipped orange-red bill and a black cap & a tern whose orange bill has a BLACK TIP (unlike arctic's all-red bill) & tern; bill: orange with black tip; cap: black \\
black-throated sparrow & a grey sparrow with a bold black throat patch and white face stripes & a sparrow with a bold BLACK THROAT and white face stripes (unlike the plain-throated chipping/field sparrows) & sparrow; throat: black patch; face: white stripes; back: grey \\
chipping sparrow & a small sparrow with a rusty crown and a black eye-line & a sparrow with a RUSTY CROWN and black eye-line (unlike the plain field sparrow) & sparrow; crown: rusty; eye-line: black; underparts: plain \\
field sparrow & a plain sparrow with a pink bill and a white eye-ring & a plain sparrow with a PINK BILL and white eye-ring, no strong marks (unlike the rusty-capped chipping sparrow) & sparrow; bill: pink; eye-ring: white; marks: minimal \\
song sparrow & a streaky brown sparrow with a central breast spot & a heavily STREAKED sparrow with a central breast SPOT (unlike the clean chipping/field sparrows) & sparrow; underparts: streaked; breast: central dark spot \\
baltimore oriole & a bright orange-and-black oriole with a black head & a BRIGHT ORANGE oriole with a fully BLACK HEAD (unlike the yellower hooded/orchard orioles) & oriole; body: bright orange; head: black; bill: straight \\
hooded oriole & a slender yellow-orange oriole with a black throat and curved bill & a YELLOW-orange oriole with a black throat and a curved bill (unlike baltimore's orange body and straight bill) & oriole; body: yellow-orange; throat: black; bill: curved \\
orchard oriole & a small dark chestnut-and-black oriole & a small DARK CHESTNUT oriole (unlike the bright-orange baltimore oriole) & oriole; body: dark chestnut and black; size: small \\
blue jay & a blue crested songbird with white and black markings & a blue crested songbird with white and black markings & songbird; color: blue; crest: yes; marks: white and black \\
cardinal & a bright red songbird with a crest and black face mask & a bright red songbird with a crest and black face mask & songbird; color: red; crest: yes; face: black mask \\
painted bunting & a multicolored bird with a blue head, green back, and red underparts & a multicolored bird with a blue head, green back, and red underparts & bird; head: blue; back: green; underparts: red \\
belted kingfisher & a stocky blue-grey bird with a shaggy crest and a big bill & a stocky blue-grey bird with a shaggy crest and a big bill & bird; build: stocky; color: blue-grey; crest: shaggy; bill: large \\
ruby-throated hummingbird & a tiny green hummingbird with a needle bill & a tiny green hummingbird with a needle bill & bird; size: tiny; color: green; bill: needle \\
scissor-tailed flycatcher & a pale flycatcher with a very long forked tail & a pale flycatcher with a very long forked tail & bird; color: pale; tail: very long and forked \\
\end{longtable}
}

\paragraph{FGVC-Aircraft.} Classes: \texttt{737-300}, \texttt{737-500}, \texttt{737-700}, \texttt{737-800}, \texttt{747-200}, \texttt{747-400}, \texttt{A319}, \texttt{A320}, \texttt{A321}, \texttt{A340-300}, \texttt{A340-600}, \texttt{MD-80}, \texttt{MD-87}, \texttt{CRJ-200}, \texttt{CRJ-700}, \texttt{Spitfire}, \texttt{Cessna 172}, \texttt{F-16A/B}, \texttt{Global Express}, \texttt{DC-3}. Descriptions are given in Table~\ref{tab:desc-fgvc}.

{\footnotesize
\setlength{\tabcolsep}{4pt}
\renewcommand{\arraystretch}{1.05}
\begin{longtable}{@{}p{0.18\linewidth}p{0.23\linewidth}p{0.30\linewidth}p{0.22\linewidth}@{}}
\caption{\textbf{FGVC-Aircraft: per-class descriptions for all three styles.} A \texttt{detailed\_all} prompt supplies the whole column; \texttt{detailed\_rare} supplies only the rows for rare classes.}\label{tab:desc-fgvc} \\
\toprule
Class & \texttt{appearance} & \texttt{contrastive} & \texttt{attribute} \\
\midrule
\endfirsthead
\multicolumn{4}{@{}l}{\itshape Table~\thetable\ continued from the previous page.}\\
\toprule
Class & \texttt{appearance} & \texttt{contrastive} & \texttt{attribute} \\
\midrule
\endhead
\midrule
\multicolumn{4}{r@{}}{\itshape Continued on the next page.}\\
\endfoot
\bottomrule
\endlastfoot
737-300 & a narrow-body Boeing 737 jet, mid-length fuselage & a Boeing 737 with a MID-length fuselage (longer than the short 737-500) & maker: Boeing; family: 737; fuselage: medium; engines: 2 underwing \\
737-500 & a narrow-body Boeing 737 jet, short fuselage & a Boeing 737 with the SHORTEST fuselage of the family (shorter than the 737-300) & maker: Boeing; family: 737; fuselage: short; engines: 2 underwing \\
737-700 & a narrow-body Boeing 737NG jet & a Boeing 737NG, SHORTER than the 737-800 & maker: Boeing; family: 737NG; fuselage: short-medium; engines: 2 underwing \\
737-800 & a narrow-body Boeing 737NG jet, long fuselage & a Boeing 737NG, the LONGEST fuselage (longer than the 737-700) & maker: Boeing; family: 737NG; fuselage: long; engines: 2 underwing \\
747-200 & a four-engine Boeing 747 jumbo jet with a short upper deck & a Boeing 747 with a SHORT upper deck and NO winglets (unlike the 747-400) & maker: Boeing; family: 747; deck: short hump; winglets: no; engines: 4 \\
747-400 & a four-engine Boeing 747 jumbo jet with winglets & a Boeing 747 with WINGLETS and an extended upper deck (unlike the 747-200) & maker: Boeing; family: 747; deck: extended hump; winglets: yes; engines: 4 \\
A319 & a short Airbus A320-family narrow-body jet & the SHORTEST Airbus A320-family jet (shorter than A320/A321) & maker: Airbus; family: A320; fuselage: short; engines: 2 underwing \\
A320 & an Airbus A320-family narrow-body jet & the MID-length Airbus A320-family jet (longer than A319, shorter than A321) & maker: Airbus; family: A320; fuselage: medium; engines: 2 underwing \\
A321 & a stretched Airbus A320-family narrow-body jet & the LONGEST, STRETCHED A320-family jet (longer than A319/A320) & maker: Airbus; family: A320; fuselage: long/stretched; engines: 2 underwing \\
A340-300 & a four-engine Airbus A340 long-haul jet & a four-engine A340, the SHORTER variant (shorter than the A340-600) & maker: Airbus; family: A340; fuselage: long; engines: 4 \\
A340-600 & a stretched four-engine Airbus A340 jet & a four-engine A340, the STRETCHED long variant (longer than the A340-300) & maker: Airbus; family: A340; fuselage: very long; engines: 4 \\
MD-80 & a twin-rear-engine McDonnell Douglas MD-80 jet & a twin-rear-engine MD jet, the LONGER standard length (longer than the MD-87) & maker: McDonnell Douglas; fuselage: long; engines: 2 rear-mounted \\
MD-87 & a shortened McDonnell Douglas MD-87 jet & a twin-rear-engine MD jet, notably SHORTENED (shorter than the MD-80) & maker: McDonnell Douglas; fuselage: shortened; engines: 2 rear-mounted \\
CRJ-200 & a small Bombardier CRJ regional jet & a small Bombardier CRJ, the SHORT regional jet (shorter than the CRJ-700) & maker: Bombardier; class: regional jet; fuselage: short; engines: 2 rear \\
CRJ-700 & a stretched Bombardier CRJ regional jet & a Bombardier CRJ, STRETCHED and larger (longer than the CRJ-200) & maker: Bombardier; class: regional jet; fuselage: stretched; engines: 2 rear \\
Spitfire & a WWII single-propeller fighter with elliptical wings & a WWII single-propeller fighter with elliptical wings & a WWII single-propeller fighter with elliptical wings \\
Cessna 172 & a small single-propeller high-wing light aircraft & a small single-propeller high-wing light aircraft & a small single-propeller high-wing light aircraft \\
F-16A/B & a single-engine jet fighter with a bubble canopy & a single-engine jet fighter with a bubble canopy & a single-engine jet fighter with a bubble canopy \\
Global Express & a large business jet & a large business jet & a large business jet \\
DC-3 & a twin-propeller 1930s airliner with a tail-wheel & a twin-propeller 1930s airliner with a tail-wheel & a twin-propeller 1930s airliner with a tail-wheel \\
\end{longtable}
}

\paragraph{Stanford Dogs.} Classes: \texttt{miniature schnauzer}, \texttt{standard schnauzer}, \texttt{giant schnauzer}, \texttt{golden retriever}, \texttt{labrador retriever}, \texttt{flat-coated retriever}, \texttt{curly-coated retriever}, \texttt{english setter}, \texttt{irish setter}, \texttt{gordon setter}, \texttt{cocker spaniel}, \texttt{sussex spaniel}, \texttt{welsh springer spaniel}, \texttt{english springer}, \texttt{pug}, \texttt{basenji}, \texttt{chihuahua}, \texttt{siberian husky}, \texttt{great dane}, \texttt{pomeranian}. Descriptions are given in Table~\ref{tab:desc-stanford}.

{\footnotesize
\setlength{\tabcolsep}{4pt}
\renewcommand{\arraystretch}{1.05}
\begin{longtable}{@{}p{0.18\linewidth}p{0.23\linewidth}p{0.30\linewidth}p{0.22\linewidth}@{}}
\caption{\textbf{Stanford Dogs: per-class descriptions for all three styles.} A \texttt{detailed\_all} prompt supplies the whole column; \texttt{detailed\_rare} supplies only the rows for rare classes.}\label{tab:desc-stanford} \\
\toprule
Class & \texttt{appearance} & \texttt{contrastive} & \texttt{attribute} \\
\midrule
\endfirsthead
\multicolumn{4}{@{}l}{\itshape Table~\thetable\ continued from the previous page.}\\
\toprule
Class & \texttt{appearance} & \texttt{contrastive} & \texttt{attribute} \\
\midrule
\endhead
\midrule
\multicolumn{4}{r@{}}{\itshape Continued on the next page.}\\
\endfoot
\bottomrule
\endlastfoot
miniature schnauzer & a small wiry-coated schnauzer with a bearded muzzle and eyebrows & the SMALLEST schnauzer (vs the medium standard and large giant) & type: schnauzer; size: small; coat: wiry; face: bearded \\
standard schnauzer & a medium wiry schnauzer with a bearded muzzle & the MEDIUM schnauzer (bigger than miniature, smaller than giant) & type: schnauzer; size: medium; coat: wiry; face: bearded \\
giant schnauzer & a large black wiry schnauzer with a bearded muzzle & the LARGEST, black schnauzer (vs the small miniature/standard) & type: schnauzer; size: large; color: black; coat: wiry \\
golden retriever & a medium dog with a dense wavy golden coat & a retriever with a WAVY GOLDEN coat (vs the short-coated labrador) & type: retriever; coat: wavy; color: golden \\
labrador retriever & a short-coated retriever, solid color, otter tail & a SHORT-coated solid-color retriever (vs the wavy golden retriever) & type: retriever; coat: short; tail: otter \\
flat-coated retriever & a black retriever with a long flat glossy coat & a BLACK LONG-FLAT-coated retriever (vs the golden's wavy gold coat) & type: retriever; coat: long flat; color: black \\
curly-coated retriever & a retriever covered in tight curls & a retriever in TIGHT CURLS all over (vs the flat/wavy coats of others) & type: retriever; coat: tight curls \\
english setter & a speckled (belton) setter with a feathered coat & a SPECKLED (belton) setter (vs solid-red irish, black-tan gordon) & type: setter; pattern: speckled belton; coat: feathered \\
irish setter & a solid chestnut-red setter with a silky coat & a SOLID CHESTNUT-RED setter (vs speckled english, black-tan gordon) & type: setter; color: solid chestnut red; coat: silky \\
gordon setter & a black-and-tan setter with a silky coat & a BLACK-AND-TAN setter (vs red irish, speckled english) & type: setter; color: black and tan; coat: silky \\
cocker spaniel & a small spaniel with long domed ears and a silky coat & the SMALL spaniel with very long domed ears (vs larger springers) & type: spaniel; size: small; ears: long domed \\
sussex spaniel & a low golden-liver spaniel with a heavy body & a LOW HEAVY golden-liver spaniel (vs the taller red-white welsh springer) & type: spaniel; build: low heavy; color: golden liver \\
welsh springer spaniel & a red-and-white medium springer spaniel & a RED-AND-WHITE springer (vs the black/liver-and-white english springer) & type: springer; color: red and white; size: medium \\
english springer & a black/liver-and-white springer spaniel & a BLACK/LIVER-AND-WHITE springer (vs the red-and-white welsh springer) & type: springer; color: black/liver and white \\
pug & a small wrinkly-faced dog with a curled tail & a small wrinkly-faced dog with a curled tail & size: small; face: wrinkled flat; tail: curled \\
basenji & a small short-coated dog with a curled tail and wrinkled forehead & a small short-coated dog with a curled tail and wrinkled forehead & size: small; coat: short; tail: curled; brow: wrinkled \\
chihuahua & a tiny dog with large erect ears & a tiny dog with large erect ears & size: tiny; ears: large erect \\
siberian husky & a wolf-like dog with erect ears and a thick coat & a wolf-like dog with erect ears and a thick coat & look: wolf-like; ears: erect; coat: thick \\
great dane & a very tall, large short-coated dog & a very tall, large short-coated dog & size: very tall/large; coat: short \\
pomeranian & a tiny fluffy dog with a fox-like face & a tiny fluffy dog with a fox-like face & size: tiny; coat: fluffy; face: fox-like \\
\end{longtable}
}

\paragraph{ImageNet-20.} Classes: \texttt{european fire salamander}, \texttt{spotted salamander}, \texttt{black and gold garden spider}, \texttt{garden spider}, \texttt{red-backed sandpiper}, \texttt{dowitcher}, \texttt{american chameleon}, \texttt{green lizard}, \texttt{grasshopper}, \texttt{cricket}, \texttt{african crocodile}, \texttt{american alligator}, \texttt{green snake}, \texttt{green mamba}, \texttt{flatworm}, \texttt{sea slug}, \texttt{dragonfly}, \texttt{damselfly}, \texttt{cabbage butterfly}, \texttt{sulphur butterfly}. Descriptions are given in Table~\ref{tab:desc-imagenet}.

{\footnotesize
\setlength{\tabcolsep}{4pt}
\renewcommand{\arraystretch}{1.05}
\begin{longtable}{@{}p{0.18\linewidth}p{0.23\linewidth}p{0.30\linewidth}p{0.22\linewidth}@{}}
\caption{\textbf{ImageNet-20: per-class descriptions for all three styles.} Rows are ordered as the ten similarity-matched pairs, frequent class first. A \texttt{detailed\_all} prompt supplies the whole column; \texttt{detailed\_rare} supplies only the rare rows.}\label{tab:desc-imagenet} \\
\toprule
Class & \texttt{appearance} & \texttt{contrastive} & \texttt{attribute} \\
\midrule
\endfirsthead
\multicolumn{4}{@{}l}{\itshape Table~\thetable\ continued from the previous page.}\\
\toprule
Class & \texttt{appearance} & \texttt{contrastive} & \texttt{attribute} \\
\midrule
\endhead
\midrule
\multicolumn{4}{r@{}}{\itshape Continued on the next page.}\\
\endfoot
\bottomrule
\endlastfoot
european fire salamander & a glossy black salamander with bold irregular yellow blotches & black salamander whose yellow markings are IRREGULAR blotches (unlike the spotted salamander's neat round spots) & salamander; ground colour: black; markings: irregular yellow blotches; skin: smooth and glossy \\
spotted salamander & a dark grey salamander with two neat rows of round yellow spots & grey salamander with two rows of ROUND yellow spots (unlike the european fire salamander's irregular blotches) & salamander; ground colour: dark grey or slate; markings: two rows of round yellow spots; skin: smooth \\
\addlinespace[2pt]
black and gold garden spider & a large orb-weaver with a boldly black-and-yellow patterned abdomen & orb-weaver with a BOLD BLACK-AND-YELLOW abdomen (unlike the garden spider's brown cross mark) & orb-weaver spider; abdomen: bold black and yellow bands; size: large; web: circular with a zigzag centre \\
garden spider & a brown orb-weaver with a pale cross-shaped mark on its abdomen & BROWN orb-weaver with a pale CROSS on the abdomen (unlike the black and gold garden spider's yellow bands) & orb-weaver spider; abdomen: mottled brown with a pale cross; size: medium; web: circular \\
\addlinespace[2pt]
red-backed sandpiper & a small shorebird with a short slightly drooping bill and a rufous back & small sandpiper with a SHORT drooping bill (unlike the dowitcher's very long straight bill) & sandpiper; size: small; bill: short, drooping at the tip; back: rufous; belly: black patch in breeding plumage \\
dowitcher & a medium shorebird with a very long straight bill probing mud & shorebird with a VERY LONG straight bill (unlike the red-backed sandpiper's short drooping bill) & shorebird; size: medium; bill: very long and straight; feeding: vertical sewing-machine probing \\
\addlinespace[2pt]
american chameleon & a slender bright green anole lizard with a pink throat fan & SLENDER green anole with a PINK throat fan (unlike the stout green lizard's blue throat) & anole lizard; colour: bright green, can turn brown; throat: pink dewlap; body: slender with a long thin tail \\
green lizard & a stout green lizard with fine dark speckling and a blue throat & STOUT green lizard with dark speckling and a BLUE throat (unlike the slender american chameleon) & lacertid lizard; colour: green with dark speckles; throat: blue in males; body: stout with a thick tail \\
\addlinespace[2pt]
grasshopper & a green or brown insect with short antennae and long jumping hind legs & insect with SHORT stout antennae (unlike the cricket's very long thread-like antennae) & orthopteran; antennae: short and stout; hind legs: long, for jumping; activity: day; wings: held roof-like \\
cricket & a brown or black insect with very long thin antennae and a flattened body & insect with VERY LONG thread-like antennae and a flattened body (unlike the grasshopper's short antennae) & orthopteran; antennae: very long and thread-like; body: flattened; activity: night; ovipositor: long in females \\
\addlinespace[2pt]
african crocodile & a large reptile with a narrow V-shaped snout and interlocking teeth showing & crocodilian with a NARROW V-shaped snout, teeth showing when closed (unlike the american alligator's broad snout) & crocodilian; snout: narrow and V-shaped; teeth: visible when jaws closed; colour: olive to grey-brown \\
american alligator & a large dark reptile with a broad rounded U-shaped snout & crocodilian with a BROAD U-shaped snout and hidden teeth (unlike the african crocodile's narrow snout) & crocodilian; snout: broad and U-shaped; teeth: hidden when jaws closed; colour: dark grey to near-black \\
\addlinespace[2pt]
green snake & a slender uniformly bright green snake with a rounded head & harmless green snake with a ROUNDED head barely wider than the neck (unlike the green mamba's coffin-shaped head) & colubrid snake; colour: uniform bright green; head: rounded, barely wider than the neck; build: very slender; harmless \\
green mamba & a bright green snake with a narrow coffin-shaped head and large eyes & green snake with a NARROW COFFIN-SHAPED head and large eyes (unlike the rounded-headed green snake) & elapid snake; colour: bright green; head: narrow and coffin-shaped; eyes: large; venomous \\
\addlinespace[2pt]
flatworm & a thin flat ribbon-like animal gliding over a surface & animal that is EXTREMELY FLAT and ribbon-like, no gills (unlike the sea slug's raised frilly back) & flatworm; body: extremely flat and leaf-like; height: negligible; appendages: none \\
sea slug & a colourful soft-bodied marine slug with frilly gills on its back & fleshy RAISED marine slug with FRILLY GILLS on its back (unlike the paper-flat flatworm) & nudibranch; body: fleshy and raised; back: exposed frilly gills and tentacles; colour: often vivid \\
\addlinespace[2pt]
cabbage butterfly & a white butterfly with black wing tips and one or two black spots & WHITE butterfly with BLACK wing TIPS (unlike the yellow sulphur butterfly) & pierid butterfly; wings: white; forewing tip: black patch; spots: one or two black on each forewing \\
sulphur butterfly & a yellow butterfly with a thin dark wing border and no black tip & YELLOW butterfly with a thin dark border and no black tip (unlike the white cabbage butterfly) & pierid butterfly; wings: yellow to orange-yellow; border: thin dark edge; forewing tip: not blackened \\
\addlinespace[2pt]
dragonfly & a large-eyed insect resting with its wings held out flat to the sides & odonate resting with wings held OUT FLAT and a thick body (unlike the damselfly's folded wings) & odonate; wings at rest: spread horizontally; body: thick; eyes: very large and touching; flight: strong \\
damselfly & a slender insect resting with its wings folded together over its back & slender odonate resting with wings FOLDED over the back (unlike the dragonfly's outspread wings) & odonate; wings at rest: folded along the body; body: very slender; eyes: separated on the sides of the head; flight: fluttering \\
\end{longtable}
}

\paragraph{PlantNet-20.} Classes: \texttt{Ophrys apifera}, \texttt{Ophrys fuciflora}, \texttt{Lamium hybridum}, \texttt{Lamium purpureum}, \texttt{Trifolium pratense}, \texttt{Trifolium medium}, \texttt{Hypericum calycinum}, \texttt{Hypericum hybrid}, \texttt{Cirsium palustre}, \texttt{Cirsium acaulon}, \texttt{Anemone apennina}, \texttt{Anemone blanda}, \texttt{Trifolium campestre}, \texttt{Trifolium dubium}, \texttt{Sedum adolphii}, \texttt{Sedum nussbaumerianum}, \texttt{Atocion rupestre}, \texttt{Moehringia ciliata}, \texttt{Nephrolepis exaltata}, \texttt{Nephrolepis biserrata}. Descriptions are given in Table~\ref{tab:desc-plantnet}.

{\footnotesize
\setlength{\tabcolsep}{4pt}
\renewcommand{\arraystretch}{1.05}
\begin{longtable}{@{}p{0.18\linewidth}p{0.23\linewidth}p{0.30\linewidth}p{0.22\linewidth}@{}}
\caption{\textbf{PlantNet-20: per-class descriptions for all three styles.} Rows are ordered as the ten similarity-matched pairs, frequent class first. A \texttt{detailed\_all} prompt supplies the whole column; \texttt{detailed\_rare} supplies only the rare rows.}\label{tab:desc-plantnet} \\
\toprule
Class & \texttt{appearance} & \texttt{contrastive} & \texttt{attribute} \\
\midrule
\endfirsthead
\multicolumn{4}{@{}l}{\itshape Table~\thetable\ continued from the previous page.}\\
\toprule
Class & \texttt{appearance} & \texttt{contrastive} & \texttt{attribute} \\
\midrule
\endhead
\midrule
\multicolumn{4}{r@{}}{\itshape Continued on the next page.}\\
\endfoot
\bottomrule
\endlastfoot
Ophrys apifera & an orchid whose furry brown lip looks like a bee, with a tucked-under yellow tip & bee orchid whose lip appendage is TUCKED UNDER and hidden (unlike Ophrys fuciflora's forward-pointing one) & genus: Ophrys; lip: brown velvety, bee-like; appendage: reflexed and hidden; sepals: pink to white \\
Ophrys fuciflora & an orchid with a broad square furry lip and a forward-pointing green appendage & bee orchid with a BROAD SQUARE lip and a FORWARD-pointing green appendage (unlike Ophrys apifera's hidden tip) & genus: Ophrys; lip: broad and squarish; appendage: forward-pointing, visible; sepals: pink, broad \\
\addlinespace[2pt]
Lamium hybridum & a small dead-nettle with pink-purple hooded flowers and deeply cut toothed leaves & dead-nettle with DEEPLY CUT sharply toothed leaves (unlike Lamium purpureum's shallowly toothed heart-shaped leaves) & genus: Lamium; flower: pink-purple hooded; leaf: deeply and sharply toothed; habit: low \\
Lamium purpureum & a small low plant with pinkish-purple hooded flowers and heart-shaped purple-tinged leaves crowded near the top & dead-nettle with HEART-SHAPED shallowly toothed purple-tinged leaves (unlike Lamium hybridum's deeply cut leaves) & genus: Lamium; flower: pink-purple hooded; leaf: heart-shaped purple-tinged; habit: low \\
\addlinespace[2pt]
Trifolium pratense & a clover with dense rounded pink-purple flower heads and pale V-marked leaflets & clover whose leaflets carry a PALE V MARK and whose head is pink-purple (unlike Trifolium medium's unmarked leaflets) & genus: Trifolium; head: dense globe, pink-purple; leaflet: oval with a pale V; habit: upright \\
Trifolium medium & a clover with deep red-purple heads and narrow leaflets lacking a pale mark & clover with NARROW UNMARKED leaflets and a deep red-purple head (unlike Trifolium pratense's V-marked leaflets) & genus: Trifolium; head: deep red-purple; leaflet: narrow, unmarked; habit: sprawling, zigzag stem \\
\addlinespace[2pt]
Hypericum calycinum & a low creeping shrub with very large solitary yellow flowers and a prominent central boss of long stamens & St John's wort with ONE VERY LARGE yellow flower and a big tuft of stamens (unlike Hypericum x's several medium flowers) & genus: Hypericum; habit: creeping shrub; flower: single very large yellow; stamens: long prominent \\
Hypericum x & a garden hybrid shrub with medium yellow flowers and arching reddish stems & garden Hypericum with SEVERAL medium yellow flowers on arching stems (unlike Hypericum calycinum's single huge bloom) & genus: Hypericum; status: garden hybrid; flower: medium yellow, several per stem; stems: arching, often reddish \\
\addlinespace[2pt]
Cirsium palustre & a tall slender spiny thistle with small dark purple heads clustered at the top & thistle that is TALL with clustered small heads on a spiny stem (unlike Cirsium acaulon's stemless ground-level head) & genus: Cirsium; height: tall and slender; head: small dark purple, in clusters; stem: spiny, winged \\
Cirsium acaulon & a thistle with almost no stem, its single purple head sitting flat in a ground-level rosette of spiny leaves & thistle that is STEMLESS, one head flat in a ground rosette (unlike the tall Cirsium palustre) & genus: Cirsium; stem: absent or very short; head: single purple at ground level; leaves: flat spiny rosette \\
\addlinespace[2pt]
Anemone apennina & a low spring flower with a pale blue many-rayed star and finely divided leaves & windflower with MANY NARROW pale blue petals in a star (unlike Anemone blanda's fewer broader petals) & genus: Anemone; flower: pale blue, many narrow petals; leaves: finely divided; habit: low woodland \\
Anemone blanda & a low spring flower with a broad blue daisy-like bloom and a paler centre & windflower with BROADER blue petals and a pale centre (unlike Anemone apennina's many narrow rays) & genus: Anemone; flower: blue with a pale centre, broader petals; leaves: divided, bronze-tinged; habit: low \\
\addlinespace[2pt]
Trifolium campestre & a small clover with rounded yellow heads that dry to pale brown papery balls & small yellow clover with LARGER heads of 20-30 flowers drying papery brown (unlike Trifolium dubium's tiny few-flowered heads) & genus: Trifolium; head: yellow, 20-30 flowers, dries brown; leaflets: nearly stalkless; habit: sprawling \\
Trifolium dubium & a tiny clover with very small yellow heads of only a few flowers & yellow clover with TINY heads of only 3-15 flowers (unlike Trifolium campestre's fuller heads) & genus: Trifolium; head: tiny yellow, 3-15 flowers; leaflet: middle one stalked; habit: prostrate \\
\addlinespace[2pt]
Sedum adolphii & a succulent with plump yellow-green leaves flushed orange-red at the tips & succulent with YELLOW-GREEN leaves merely red-TIPPED (unlike Sedum nussbaumerianum's wholly copper-orange leaves) & genus: Sedum; leaves: plump, keeled, yellow-green with red tips; habit: branching succulent rosettes \\
Sedum nussbaumerianum & a succulent with thick copper-orange leaves in loose open rosettes & succulent whose leaves are COPPER-ORANGE throughout (unlike Sedum adolphii's yellow-green leaves with red tips) & genus: Sedum; leaves: thick, copper to orange; rosettes: loose and open; habit: sprawling succulent \\
\addlinespace[2pt]
Atocion rupestre & a small rock plant with pink-white notched petals on slender forking stems & small rock plant with NOTCHED pink-white petals on forking stems (unlike Moehringia ciliata's unnotched white petals) & family: Caryophyllaceae; petals: 5, notched, pink to white; leaves: grey-green paired; habit: tufted on rock \\
Moehringia ciliata & a tiny cushion plant with small pure white petals and narrow fringed leaves & cushion plant with UNNOTCHED pure white petals and hair-fringed leaves (unlike Atocion rupestre's notched petals) & family: Caryophyllaceae; petals: 5, white, not notched; leaves: narrow, hair-fringed; habit: dense cushion \\
\addlinespace[2pt]
Nephrolepis exaltata & a fern with arching bright green fronds of many small close-set leaflets & fern with SHORT CLOSE-SET wavy leaflets on arching fronds (unlike Nephrolepis biserrata's long spaced toothed leaflets) & genus: Nephrolepis; frond: arching, to 1 m; pinnae: short, close-set, wavy-edged; habit: dense clump \\
Nephrolepis biserrata & a large fern with long stiff fronds of widely spaced toothed leaflets & fern with LONG WIDELY SPACED toothed leaflets on stiff fronds (unlike Nephrolepis exaltata's short close-set ones) & genus: Nephrolepis; frond: long and stiff, to 2 m; pinnae: long, spaced, toothed; habit: coarse open clump \\
\end{longtable}
}

\end{document}